\documentclass[runningheads]{llncs}

\usepackage{eccv}

\usepackage{eccvabbrv}

\usepackage{graphicx}
\usepackage{booktabs}
\usepackage{float}
\usepackage{cuted}%%\strip
\usepackage{wrapfig}
\usepackage{algpseudocode}
\usepackage{amssymb}
\usepackage{times}
\usepackage{epsfig}
\usepackage{graphicx}
\usepackage{amsmath}
\usepackage{amssymb}
\usepackage{booktabs}
\usepackage{algorithm}
\usepackage{algpseudocode}
\usepackage{xcolor}
\usepackage{multirow}
\usepackage[table]{xcolor}
\usepackage{bbding}

\usepackage[accsupp]{axessibility}  % Improves PDF readability for those with disabilities.

\usepackage[urlcolor=red]{hyperref}

\usepackage{orcidlink}

\begin{document}

\title{From Draft to Draft-Free: One-Step Video Object Removal via Privileged Distillation and Fast Planting} 

\titlerunning{From Draft to Draft-Free}

% \author{Zizhao Chen\inst{1} \and
% Ping Wei\inst{1} \and Guang Dai\inst{2} \and Jingdong Wang\inst{4} \and Mengmeng Wang\inst{2,3}
% }

% % TODO FINAL: Replace with an abbreviated list of authors.
% % \authorrunning{F.~Author et al.}

% % TODO FINAL: Replace with your institution list.
% \institute{State Key Laboratory of Human-Machine Hybrid Augmented Intelligence, \\
% Institute of Artificial Intelligence and Robotics, Xi’an Jiaotong University \and
% SGIT AI Lab, State Grid Corporation of China \and Zhejiang University of Technology \and
% Baidu\\

\author{
Zizhao Chen\inst{1} \and
Ping Wei \inst{1}\textsuperscript{\dag} \and
Guang Dai\inst{2} \and
Jingdong Wang\inst{4} \and
Mengmeng Wang\inst{3,2}\textsuperscript{\dag}
}
\authorrunning{Z. Chen et al.}
\institute{
State Key Laboratory of Human-Machine Hybrid Augmented Intelligence,\\
Institute of Artificial Intelligence and Robotics, Xi'an Jiaotong University \and
SGIT AI Lab, State Grid Corporation of China \and
Zhejiang University of Technology \and
Baidu \\
\texttt{\url{https://github.com/bigD233/D2DF}
}
}

\maketitle

\renewcommand{\thefootnote}{\dag}
\footnotetext{Corresponding authors. Contact: \texttt{pingwei@xjtu.edu.cn}, \texttt{mengmewang@gmail.com}.}
\renewcommand{\thefootnote}{\arabic{footnote}}

\begin{figure*}
    \vspace{-0.5cm}
    \centering
    \includegraphics[width=\textwidth]{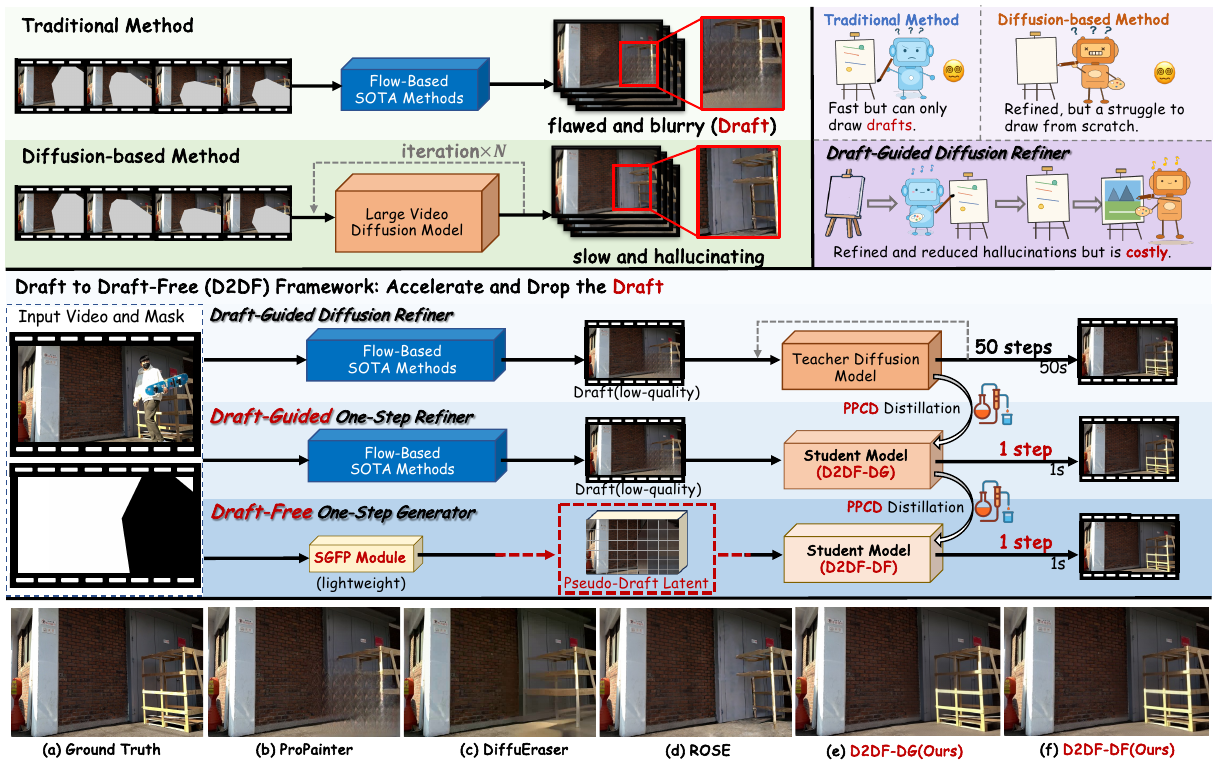}
    \caption{Overview of our three-staged \textbf{D2DF} framework. To address the limitations of traditional and diffusion-based methods, we combine them as a refiner. However, this incurs significant time costs and depends on external drafts. Through PPCD distillation and the SGFP module, we obtain the one-step video object removal models D2DF-DG and D2DF-DF. Visual comparisons from (a) to (f) demonstrate that D2DF exhibits powerful background reconstruction regardless of whether drafts are used as prior information.}
    \vspace{-0.8cm}
    \label{fig:teaser}
\end{figure*}

\begin{abstract}
  Video object removal is a fundamental yet challenging task in video editing. Despite recent progress, existing methods typically fall into two categories. Traditional approaches based on optical flow or attention mechanisms often introduce noticeable artifacts and yield unnatural results. In contrast, diffusion-based methods improve visual realism but demand multiple denoising steps, limiting their practicality.
To address these issues, we propose From-Draft-to-Draft-Free (\textbf{D2DF}), a framework that distills the ability of transforming coarse drafts into refined videos into a one-step video generation model. Within D2DF, a teacher model is trained to refine low-quality removal results (“drafts”) into high-fidelity videos by multiple steps. Then, through Prior-Privileged Consistency Distillation (\textbf{PPCD}), we distill this capability into a student model that performs one-step removal conditioned on the draft. To eliminate draft dependency, we introduce a Self-Guided Fast Planting (\textbf{SGFP}) module based on our Temporal Masked Transformer that autonomously generates scene-consistent pseudo-drafts in latent space, enabling a fully \textbf{draft-free} one-step model.
Extensive experiments show that both draft-conditioned and draft-free versions achieve state-of-the-art performance on multiple metrics, surpassing traditional and multi-step generative methods in both quality and efficiency. The denoising process for a single video takes only about 1 second.
\end{abstract}

\section{Introduction}
\label{sec:intro}

Video object removal (VOR) aims to remove undesired objects from videos and reconstruct the background in the occluded regions. As an essential task in video inpainting\cite{ebdelli2015video,wang2023unsupervised,zhang2020autoremover}, it has a wide range of applications.
The challenge is intrinsically difficult, often requiring globally consistent spatial–temporal content generation\cite{tang2011video,bertalmio2001navier,ding2025homogen} for regions that cannot be propagated from adjacent frames.

Traditional VOR methods rely on optical-flow propagation or shallow transformer architectures~\cite{zhou2023propainter,li2022towards,zhang2022flow,gao2020flow,liu2021fuseformer,zeng2020learning}. While they propagate background textures, their limited generative capacity yields structural distortions and blurry reconstructions, especially for large occlusions (Fig.~\ref{fig:teaser}(b)). This ``draft" output contains noticeable flaws.

The emergence of large video diffusion models~\cite{rombach2022high,yang2024cogvideox,peebles2023scalable,miao2025rose,li2025diffueraser,gu2024coherent} offers
a path to solving this ``generative capacity" bottleneck, significantly enhancing realism. However, two critical problems remain. First, \textbf{Hallucination}:  weak priors cause diffusion models to ``hallucinate"—generating plausible but semantically incorrect content (Fig.~\ref{fig:teaser}(c)(d)), failing deterministic reconstruction. Second, \textbf{Slowness}: diffusion inference is intrinsically slow (tens of DDIM steps~\cite{song2020denoising}), hindering practical deployment.

To solve the hallucination issue, we observe that flow-based methods efficiently propagate reliable pixel information across frames. This process is fast and fills parts of the masked region with valid content. While their ``draft" output can be blurry (Fig. \ref{fig:teaser}(b)), it provides a strong, partially reconstructed starting point. We use this ``draft" to guide a diffusion ``refiner" to avoid hallucination. To solve the slowness issue, we argue that the full multi-step refinement is unnecessary given this strong ``draft" condition. The task is simplified enough that the ``refiner" can be distilled into a single, efficient step. Our ultimate goal, however, is an autonomous model that learns this paradigm but discards the external ``draft" prior at inference. 

Based on this insight, we propose the ``From Draft to Draft-Free'' framework \textbf{(D2DF)}, which progressively derives a draft-guided one-step refiner (D2DF-DG) and a fully draft-free one-step generator (D2DF-DF). Our framework (Fig.~\ref{fig:teaser}) has three stages:

\noindent \textbf{Stage I: The Draft-Guided Diffusion Teacher.}
We first train a multi-step diffusion teacher (D-LDM) conditioned on external ``drafts" (e.g., ProPainter~\cite{zhou2023propainter}). The teacher refines the flawed draft into a high-quality object removal video through 50 steps.

\begin{figure}[!t]
    \centering
    \includegraphics[width=\linewidth]{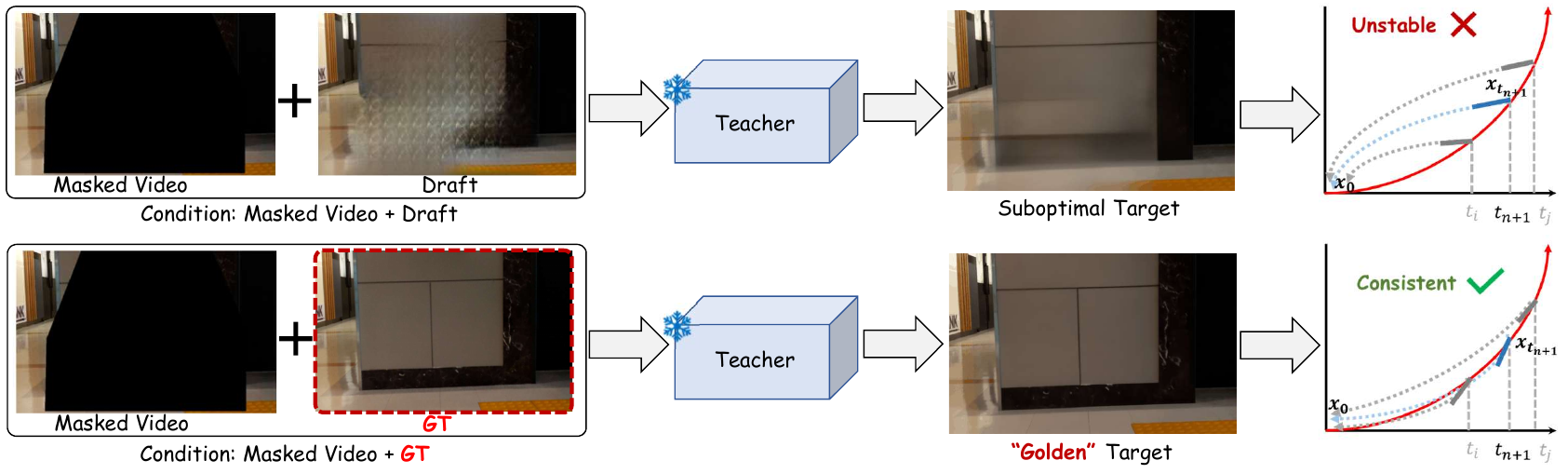}
    \caption{Comparison of D-LDM outputs under different conditions. The middle column shows the results obtained by 4-step DDIM denoising of the teacher network. The solid red line represents the ground-truth denoising trajectory. The dashed line represents the solution trajectory of the teacher. The short solid line indicates the target provided by the teacher at the current timestep.}
    \vspace{-0.5cm}
    \label{fig:ppcd_compare}
\end{figure}

\noindent \textbf{Stage \uppercase\expandafter{\romannumeral2}: The Draft-Guided One-Step Refiner} (D2DF-DG). Our goal is to distill the 50-step teacher into a 1-step student (D2DF-DG). A naive consistency distillation \cite{song2023consistency,luo2023latent,song2023improved} would fail for the reason that when conditioned on the flawed draft, the teacher must reconstruct from varying levels of damage. This forces it onto an unstable solution trajectory that significantly deviates from the ideal path to the ground truth (GT). As Fig.\ref{fig:ppcd_compare} illustrates, this noisy path provides an inconsistent and Suboptimal Target for the student, making convergence difficult. To solve this, we introduce \textbf{Prior-Privileged Consistency Distillation (PPCD)}. We replace the flawed draft with the GT as the condition of the teacher. This ``truth-injected" strategy compels the teacher to generate a perfectly consistent and stable trajectory. This ideal path, which stays aligned with the GT, provides a ``Golden'' Target for the student, dramatically improving distillation effectiveness. Crucially, the student (D2DF-DG) still accepts the draft as input, but now learns a clean, one-step mapping to this ``golden" output.

\noindent \textbf{Stage \uppercase\expandafter{\romannumeral3}: The Draft-Free One-Step Generator} (D2DF-DF). Despite the impressive performance of D2DF-DG, it still relies on external drafts. To eliminate this dependency, we design the \textbf{Self-Guided Fast Planting (SGFP)} module, a lightweight network based on Temporal Masked Transformer (TMT) that constructs a ``pseudo-draft" in the latent space. This differs fundamentally from drafts in explicit pixel-space from typical methods like ProPainter\cite{zhou2023propainter}. We then use PPCD again to distill D2DF-DG (as teacher) into our final student, D2DF-DF. This model inherits the one-step speed but is now fully autonomous, achieving end-to-end generation without external priors.

Our D2DF-DF achieves superior performance on three benchmarks~\cite{sagong2022rord,wu2024towards,yoon2024raccoon}, exhibiting strong robustness, especially under large-mask scenarios. Crucially, our models set a new standard for efficiency: D2DF-DF achieves state-of-the-art results while being over 40 times faster than competing SOTA diffusion methods like ROSE\cite{miao2025rose}. This combination of quality, speed, and autonomy represents a significant step forward for practical video editing.

Our contributions can be summarized as follows:

\noindent $\bullet$ We propose a novel From-Draft-to-Draft-Free (D2DF) framework, a progressive methodology to decouple latency and dependency, producing two models: a high-efficiency one-step refiner (D2DF-DG) and a fully autonomous one-step generator (D2DF-DF).

\noindent $\bullet$ In D2DF, we introduce Prior-Privileged Consistency Distillation (PPCD), a novel ``truth-injected" distillation where a teacher uses GT to create a golden target.

\noindent $\bullet$ We design the Self-Guided Fast Planting (SGFP) module, a lightweight latent-space prior generator, combined with PPCD, to create a fully autonomous, end-to-end, one-step model.

\section{Related Work}
\label{sec:rw}

\subsection{Diffusion-based Video Generation}
\vspace{-2pt}
Diffusion models have recently achieved remarkable success in visual generation by iteratively denoising noisy inputs \cite{blattmann2023align,dhariwal2021diffusion,jin2025flovd,podell2023sdxl,wang2024microcinema,weng2024art,xing2024simda}. Following advances in text-to-image diffusion, recent works extend this paradigm to video generation. Text-to-video (T2V) models \cite{ho2022imagen,blattmann2023stable,wang2023videocomposer,wei2024dreamvideo,yang2024cogvideox,wu2023tune} integrate temporal modules into 2D UNet architectures, while image-to-video (I2V) methods such as DynamiCrafter \cite{xing2024dynamicrafter} and PixelDance \cite{zeng2024make} demonstrate strong motion and scene modeling ability. More recently, Transformer-based diffusion models (e.g., DiT, SORA-like \cite{brooks2024video,peebles2023scalable}) further improve temporal reasoning and visual fidelity.
% Diffusion-based video inpainting methods leverage visible pixels or motion priors as conditions to fill occluded regions, achieving realistic and coherent reconstructions. However, they still require tens of denoising steps, resulting in a high inference cost.

\vspace{-10pt}
\subsection{Video Inpainting}
\vspace{-2pt}
Most existing video inpainting approaches adopt a two-stage framework.
In the first stage, prior information is extracted from the input video, such as motion cues \cite{kim2019deep,tang2011video,xu2019deep}, homography transformations\cite{lee2019copy} or low-resolution reconstructions \cite{wang2019video}.
In the second stage, based on these priors, the missing regions are filled by a carefully designed generative network. Typical architectures include image inpainting models \cite{gao2020flow,xu2019deep,zhang2022inertia}, 3D convolutional networks \cite{kim2019deep}, and Transformer-based frameworks \cite{zhang2022flow}.
Recent studies have combined the advantages with the spatio-temporal modeling power of video Transformers \cite{zhou2023propainter,li2022towards,zhang2022flow,gao2020flow}, leading to a series of two-stage hybrid methods.
Recently, researchers have introduced video diffusion models for video inpainting \cite{li2025diffueraser,gu2024advanced,bian2025videopainter,miao2025rose,lee2025video,ding2025homogen}. These models substantially improve the realism and coherence of the generated videos, yet they still suffer from hallucination in the reconstructed regions.
To address these issues, our work introduces the SGFP module to provide strong priors, together with a Prior-Privileged Consistency Distillation (PPCD) framework that enables one-step video generation, effectively balancing quality and efficiency.

\vspace{-10pt}
\subsection{Consistency Model}
\vspace{-2pt}
Diffusion models achieve impressive generative performance but suffer from slow inference due to their multi-step sampling process. To accelerate generation, several strategies have been proposed, such as improved ODE solvers \cite{lu2022dpm}, neural operators \cite{zheng2023fast}, and model distillation \cite{meng2023distillation,salimans2022progressive,sauer2024adversarial}. Among them, consistency distillation \cite{song2023consistency} has emerged as an effective solution. Instead of relying on iterative denoising, it trains a student consistency model to directly predict clean samples from intermediate noisy states, enabling one-step generation while preserving sample quality. Subsequent works, such as Latent Consistency Models (LCM) \cite{luo2023latent,luo2023lcm}, extend this paradigm into latent space for higher efficiency and lower memory cost. Recent improvements \cite{lu2024simplifying,song2023improved,geng2024consistency} further enhance stability and fidelity, making consistency distillation a promising framework for efficient generative modeling.
However, consistency models still face challenges in adapting to task-specific applications, especially for video generation. To address this, we propose Prior-Privileged Consistency Distillation (PPCD), which provides more stable and consistent solution trajectories for video generation tasks.

\section{Method}

\subsection{Preliminaries}
\label{sec:preliminaries}

Latent Diffusion Models (LDM)\cite{rombach2022high} establish the generative modeling process in the latent space. The training objective matches the predicted denoised sample $\mathbf{x}_0$ with the ground truth through a simplified variational bound:
\begin{equation}
    \mathcal{L}_{LDM} = \mathbb{E}_{x, t, \epsilon \sim \mathcal{N}(0,\mathbf{I})} 
    \left[\|\boldsymbol{\epsilon} - \boldsymbol{\epsilon}_\theta(\mathbf{x}_t, t)\|_2^2\right],
    \label{eq:ddpm_loss}
\end{equation}
where $\mathbf{x}_t$ is the noisy input obtained through the forward diffusion process with timestep $t$, and $\boldsymbol{\epsilon}_\theta$ is a neural network that predicts the noise component. This objective effectively establishes a progressive denoising mapping that transforms pure noise $\mathbf{x}_T$ into clean data $\mathbf{x}_0$ through a Markov chain of $T$ steps. This mapping can be formulated as a parameterized reverse process $\mathbf{p}_\theta:(\mathbf{x}_t,t) \longmapsto \mathbf{x}_{t-1}$. However, this requires an iterative denoising process.

Consistency Model \cite{song2023consistency} attempts to establish mappings through a consistency function, defined as $\mathbf{f}:(\mathbf{x}_t,t) \longmapsto \mathbf{x}_\delta$, where $\delta$ is a fixed small positive number. Its consistency is reflected in one of the key properties of this function: 
\begin{equation}
    \mathbf{f}(\mathbf{x}_t,t)=\mathbf{f}(\mathbf{x}_{t'},t'),\quad \forall t,t'\in[\epsilon,T].
    \label{eq:self_consistency}
\end{equation}
Ideally, the consistency function established through learnable neural networks $f_\theta$ enables the consistency model to recover the original data $\mathbf{x}_0$ from the initial distribution $\mathbf{x}_T \sim \mathcal{N}(\mathbf{0}, T^2 \boldsymbol{I})$ in one step of denoising. Learnable parameters $\theta$ are trained via Consistency Distillation \cite{song2023consistency,song2023improved,geng2024consistency}. The distillation goal is to minimize the output discrepancy between random adjacent data points, thereby ensuring self-consistency in the sense of probability. Formally, the consistency loss is defined as follows:
\begin{small}
\begin{equation}
    \mathcal{L}(\mathbf{\theta}, \mathbf{\theta}^-;\Phi)=\mathbb{E}_{\boldsymbol{x},t} 
    \left[d\left(\mathbf{f}_\mathbf{\theta}(\mathbf{x}_{t_{n+1}},t_{n+1}),\mathbf{f}_{\mathbf{\theta}^{-}}(\hat{\mathbf{x}}^\phi_{t_n},t_n)\right)\right],
    \label{eq:consistency_loss}
\end{equation}
\end{small}
\normalsize where $0 = t_1 < t_2 < \cdots < t_N = T$, and $n$ is uniformly distributed over the set $\{1, 2, \cdots, N - 1\}$. 
$\Phi(\cdot)$ denotes the one-step ODE \cite{lu2022dpm} solver applied to PF-ODE\cite{song2020score}, the model parameter $\mathbf{\theta^-}$ is obtained from the exponential moving average (EMA) of $\mathbf{\theta}$, and $d(\cdot,\cdot)$ is a chosen metric function for measuring the distance between two samples. $\hat{x}^{\phi}_{t_n}$ is determined using the equation 
$\hat{\mathbf{x}}^{\phi}_{t_n} := \Phi(\mathbf{x}_{t_{n+1}}, t_{n+1}, t_n; \phi)$. $\hat{\mathbf{x}}^{\phi}_{t_n}$ is estimated by a teacher model with an ODE solver in consistency distillation.

\begin{figure}[!t]
    \centering
    \includegraphics[width=\linewidth]{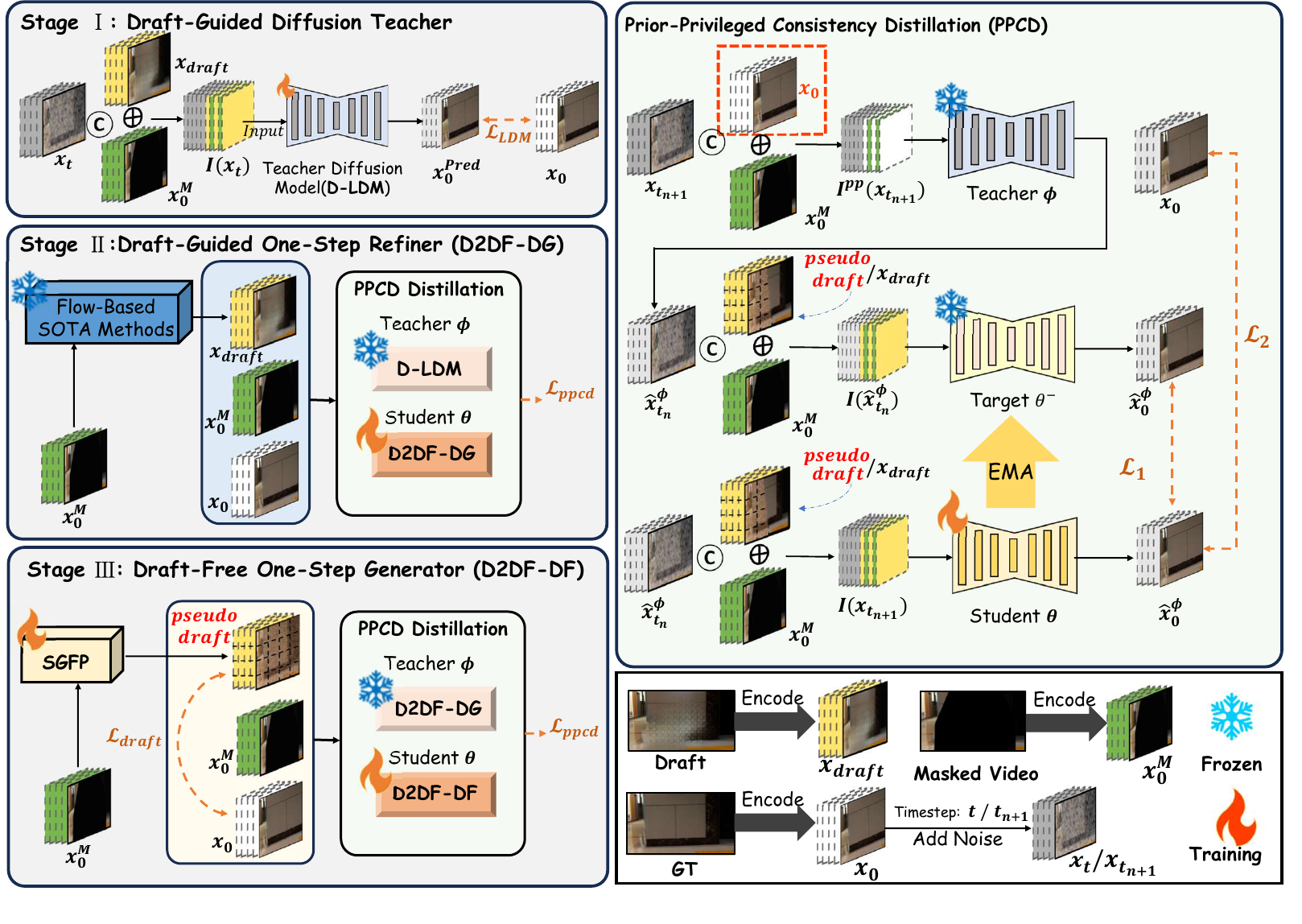}
    \caption{Pipeline of D2DF. On the left are the three stages of our framework, while on the right is the PPCD distillation we propose. }
    \vspace{-0.5cm}
    \label{fig:framework}
\end{figure}

\subsection{Stage \uppercase\expandafter{\romannumeral1}: Draft-Guided Diffusion Teacher}
\label{sec:stage1}
Our first stage simplifies the task into a refinement process. We introduce a strong conditional prior (\textbf{draft}) from existing methods (e.g., ProPainter \cite{zhou2023propainter}). This process establishes a sufficiently powerful teacher network to provide a mapping from draft to refined video in multiple steps. Thus, we train a Draft-Guided Diffusion Teacher model (D-LDM), denoted as $\phi$. This model is conditioned on both the draft latent $\mathbf{x}_\text{draft}$ and the masked video latent $\mathbf{x}_{0}^M$. At timestep $t$, the model input $I(\mathbf{x}_t)$ is defined as the concatenation of the noised input $\mathbf{x}_t$ and the conditional information:
\begin{equation}
    I(\mathbf{x}_t)= Cat(\mathbf{x}_t,\mathbf{x}_0^{M}+\mathbf{x}_{\text{draft}}),
\end{equation}
where $\mathbf{x}_t$ denotes the noised input through the forward diffusion process and $Cat(\cdot)$ denotes the latents concatenation.

The D-LDM model is then trained by substituting this conditional input $I(\mathbf{x}_t)$ for $\mathbf{x}_t$ in the LDM training objective (Eq.\ref{eq:ddpm_loss}). The objective for the teacher network $\boldsymbol{\epsilon}_\phi$ becomes:
\begin{equation}
    \mathcal{L}_{D-LDM} = \mathbb{E}_{x, t, \epsilon \sim \mathcal{N}(0,\mathbf{I})} 
    \left[\|\boldsymbol{\epsilon} - \boldsymbol{\epsilon}_\phi(I(\mathbf{x}_t), t)\|_2^2\right].
\end{equation}
This strong draft conditioning significantly reduces the mapping complexity, making a one-step approximation feasible. The resulting D-LDM $\phi$ achieves strong generative performance via iterative (e.g., 50-step) DDIM sampling.

\subsection{Stage \uppercase\expandafter{\romannumeral2}: Prior-Privileged Consistency Distillation for Draft-Guided One-Step Refiner}
\label{sec:stage2}

In the second stage, we distill the D-LDM teacher model ($\phi$) into a one-step, draft-guided student model, \textbf{D2DF-DG}. 
While the theory of consistency distillation (CD) is promising, it suffers from high sampling variance from the teacher model, especially when conditioned on imperfect drafts, which destabilizes the distillation process.

To overcome this, we introduce Prior-Privileged Consistency Distillation (PPCD). Standard distillation from a teacher, conditioned on flawed drafts, leads to unstable and suboptimal trajectories. Our key idea is to inject the ground-truth video as a ``privileged" prior into the teacher model $\phi$ during distillation only. This provides an ideal, ``golden" target trajectory and ensures the mapping is anchored in a convergent solve space.

Formally, the teacher model serves as a trajectory provider. To ensure it provides a stable and reliable consistency trajectory, we replace the draft latent $\mathbf{x}_{\text{draft}}$ with the ground-truth latent $\mathbf{x}_0$ as shown in Fig.\ref{fig:framework}(PPCD). The teacher's privileged input at timestep $t$ becomes:
\begin{equation}
\mathbf{I}^{pp}(\mathbf{x}_t) = Cat(\mathbf{x}_t, \mathbf{x}_0^{M} + \mathbf{x}_0).
\end{equation}
Given this privileged input, the teacher network produces a denoised latent $\hat{\mathbf{x}}^{\phi}_{t_n} = \Phi(I^{pp}(\mathbf{x}_{t_{n+1}}), t_{n+1}, t_n; \phi)$ at the next timestep along the diffusion trajectory.

Based on the principle of consistency distillation (Eq.\ref{eq:consistency_loss}), the student network $\mathbf{f}_{\mathbf{\theta}}$ (and its EMA target $\mathbf{f}_{\mathbf{\theta}^{-}}$) is trained to match the consistency property. However, the student still operates using the \textbf{draft} as input, as it will at inference time. We obtain the target latent and prediction latent:
\begin{align}
\hat{\mathbf{x}}^{\phi}_{0} & = \mathbf{f}_{\mathbf{\theta}^{-}}(Cat(\hat{\mathbf{x}}^\phi_{t_n},\mathbf{x}_0^M + \mathbf{x}_{\text{draft}}),t_n), \\
\mathbf{x}^{pred}_{0} & = \mathbf{f}_{\mathbf{\theta}}(Cat(\hat{\mathbf{x}}^\phi_{t_{n+1}},\mathbf{x}_0^M + \mathbf{x}_{\text{draft}}),t_{n+1}),
\end{align}
where $\mathbf{f}_{\mathbf{\theta}^{-}}$ denotes the target model and $\theta^-$ is an EMA version of the student network $\theta$. The meanings of $t_n$ and $t_{n+1}$ are the same as in Eq.\ref{eq:consistency_loss}.
Ultimately, our PPCD utilizes the following objectives to optimize the student network $\theta$:
\begin{small}
\begin{equation}
\mathcal{L}_{\text{ppcd}} =  \mathbb{E}_{\boldsymbol{x},t} 
    \left[d\left(\mathbf{x}^{pred}_{0},\hat{\mathbf{x}}^{\phi}_{0}\right)\right] + \alpha \mathbb{E}_{\boldsymbol{x},t} 
    \left[d\left(\mathbf{x}^{pred}_{0},\mathbf{x}_{0}\right)\right],
\end{equation}
\end{small}
where $d(\cdot,\cdot)$ is the mean square error of two samples and $\alpha$ is a hyperparameter. The first term enforces consistency with the privileged trajectory, while the second term adds a direct ground-truth constraint.

Using the trained D-LDM as the teacher, we apply PPCD to obtain \textbf{D2DF-DG}, a one-step, draft-conditioned object removal refiner model.

\subsection{Stage \uppercase\expandafter{\romannumeral3}: Self-Guided Fast Planting (SGFP) for Draft-Free One-Step Generator }
\label{sec:stage3}

The D2DF-DG model from Stage \uppercase\expandafter{\romannumeral2} is fast but still depends on drafts generated by external methods, incurring inference overhead. In our final stage, we eliminate this dependency by proposing the \textbf{Self-Guided Fast Planting (SGFP)} module, resulting in the final draft-free model, \textbf{D2DF-DF}.

\begin{wrapfigure}{r}{0.5\textwidth}
    \centering
    \includegraphics[width=\linewidth]{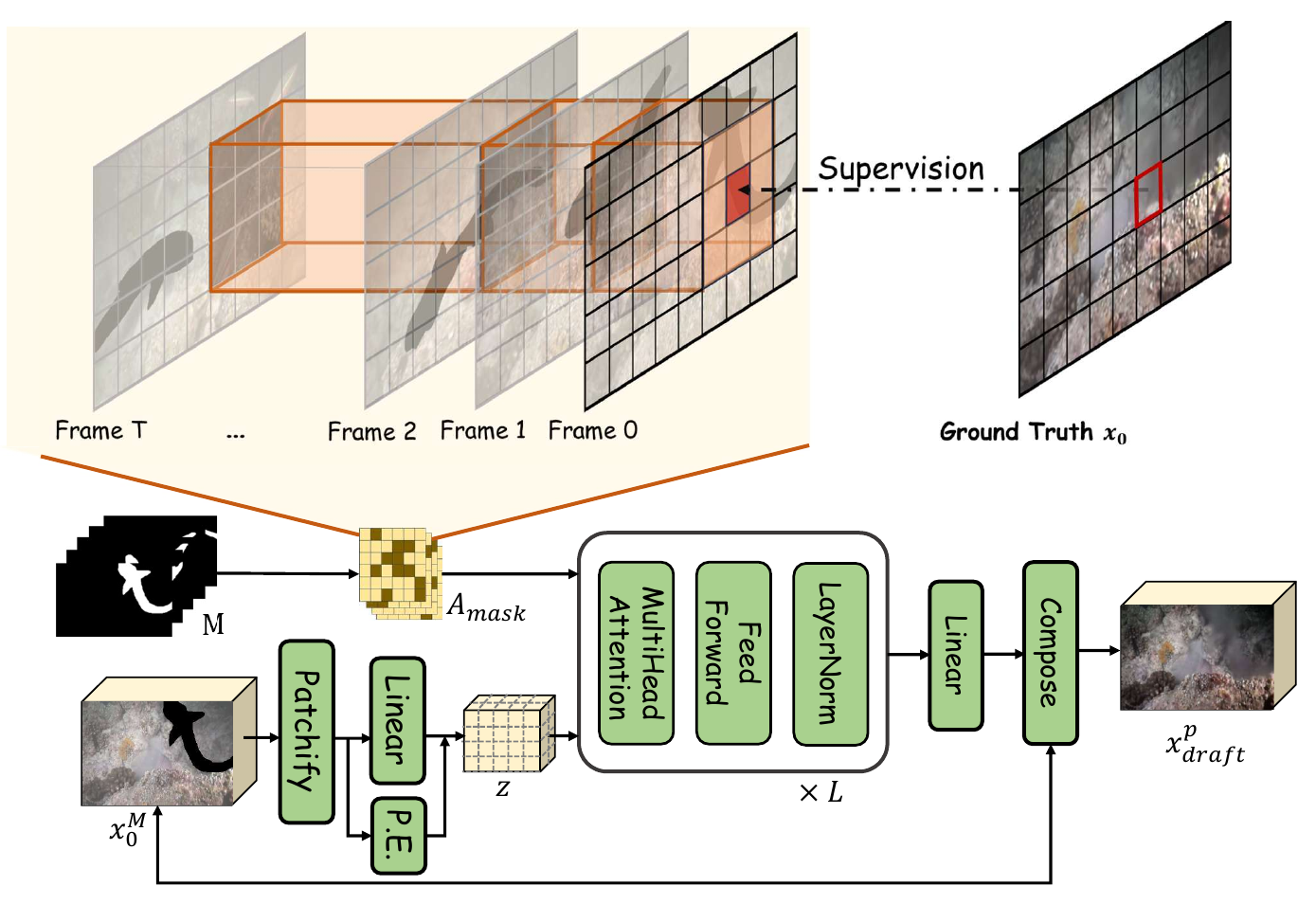}
    \vspace{-20pt}
     \caption{Pipeline of SGFP. The entire reconstruction operation is in the latent space, constrained by the attention mask based on valid patches and spatial windows. }
    \vspace{-20pt}
    \label{fig:sgfp}
    % \vspace{10pt}
\end{wrapfigure}

The SGFP module (Fig.\ref{fig:sgfp}) learns to reconstruct a ``pseudo-draft" directly in latent space using spatio-temporal information from unoccluded regions. The module is built upon our Temporal Masked Transformer (TMT) architecture, which aims to rapidly reconstruct masked regions. Given a masked draft latent $\mathbf{x}_{0}^M \in \mathbb{R}^{C \times T \times H \times W}$ and a binary mask $\mathbf{M} \in \{0, 1\}^{1 \times T \times H \times W}$ (where 0 denotes masked regions and 1 denotes known regions), the objective of SGFP is to get a completed latent tensor $\mathbf{x}_{recon}$.

To handle the high-dimensional video latents while maintaining computational tractability, we first partition the spatial dimensions $(H, W)$ into a series of non-overlapping \textit{patches} of size $P \times P$. This reshapes the input into a sequence of $N = T \times (H/P) \times (W/P)$ tokens, where each token is a vector of dimension $C \cdot P^2$. Each patch vector $\mathbf{p}_i$ is then embedded into a lower-dimensional space $\mathbb{R}^{D_e}$ using a linear layer, combined with a positional embedding:
$
\mathbf{z}_i = \text{Linear}(\mathbf{p}_i) + \mathbf{e}_{pos}(i) ,
$
where $D_e$ is the embedding dimension and $\mathbf{z}_i$ is the input to the TMT layers. The term $\mathbf{e}_{pos}(i)$ is a learnable positional encoding generated by a small MLP from the patch's normalized 3D coordinates $(t, h, w)$, providing spatiotemporal awareness. 

% \begin{figure}[!t]
%     \centering
%     \includegraphics[width=\linewidth]{figures/SGFP1.pdf}
%     \caption{Pipeline of SGFP. The entire reconstruction operation is in the latent space, constrained by the attention mask based on valid patches and spatial windows. }
%     % \vspace{-0.5cm}
%     \label{fig:sgfp}
% \end{figure}

% \vspace{-0.5cm}
The key to the ``self-guided'' nature of SGFP lies in the construction of its self-attention mechanism. The computation must rely solely on known, unmasked information. We first compute a mask ratio $r_i = \text{Mean}(\mathbf{M}_{patch, i})$ for each patch. If $r_i$ exceeds a threshold $\tau$, the patch is considered ``invalid''. In the self-attention calculation, all patches $\mathbf{z}_i$ can act as Queries, but only ``valid'' patches (i.e., $r_j \le \tau$) are permitted to serve as Keys and Values. This is enforced via an attention mask $\mathbf{A}_{mask}$:
$$ 
\mathbf{A}_{mask}(i, j) = \begin{cases} 0 & \text{if } r_j \le \tau \text{ and } \text{dist}(i, j) \le R \\ -\infty & \text{otherwise} \end{cases},
$$
where $\text{dist}(i, j) \le R$ is an optional spatial window constraint that further restricts attention to a $(2R+1) \times (2R+1)$ spatial neighborhood of the query patch $i$ (across all time frames $T$), reducing complexity and reinforcing local priors. After $L$ Transformer layers, the output sequence $\mathbf{z}_i^{(L)}$ is projected back to the original patch dimension $\mathbb{R}^{C \cdot P^2}$ and ``un-folded'' to reconstruct the latent variable $\mathbf{x}_{recon} \in \mathbb{R}^{C \times T \times H \times W}$. Finally, the pseudo-draft $\mathbf{x}_{\text{draft}}^{p}$ is generated by composing the reconstruction and the original input using the mask $\mathbf{M}$, ensuring the fidelity of unmasked regions:
\begin{equation}
   \mathbf{x}_{\text{draft}}^p = \mathbf{x}_{recon} \odot (1 - \mathbf{M}) + \mathbf{x}_{0}^M \odot \mathbf{M} . 
\end{equation}

We then treat the D2DF-DG model (from Stage \uppercase\expandafter{\romannumeral2}) as the new teacher and employ the PPCD framework again to distill the final draft-free student, \textbf{D2DF-DF}, which now incorporates the SGFP module. We trained the SGFP together with the subsequent generation module, incorporating constraints for pseudo-drafts:
\begin{equation}
\mathcal{L}_{\text{draft}} =  \mathbb{E}_{\boldsymbol{x},t} 
    \left[d\left(\mathbf{x}^{p}_{draft},\mathbf{x}_{0}\right)\right].
\end{equation}
By introducing SGFP, we obtain D2DF-DF, an end-to-end, draft-free video object removal model. The entire progressive pipeline is summarized in Figure \ref{fig:framework}.

\section{Experiments}

\subsection{Experiment Setup}
\noindent \textbf{Implementation Details.}
We employ CogVideoX-5B-I2V\cite{yang2024cogvideox} (480$\times$720 resolution) as the pre-trained weights for training the first-stage D-LDM network. The D-LDM model is trained with a learning rate of $5\times10^{-5}$ and uses 50 DDIM steps for inference. The ``draft" is obtained using the ProPainter\cite{zhou2023propainter}.
In the PPCD distillation, the hyperparameter $\alpha$ is set to 1, with a batch size of 4 and a learning rate of $2\times10^{-5}$. Sampling timesteps $t_1,t_2,...,t_N$ are performed uniformly at 50-step intervals over 1k timesteps. The EMA decay rate of the target network is set to 0.99. 
For the SGFP module, the number of layers $L$ was set to 4, the spatial radius $R$ to 2, the patch size $P$ to 10, the embedding dimension $D_e$ to 512, and the mask threshold $\tau$ to 0.7. In the third-stage training, the ratio of $\mathcal{L}_{\text{draft}}$ to $\mathcal{L}_{\text{ppcd}}$ is 1:5, and the learning rate for the SGFP module parameters is $1\times 10^{-4}$. We train only the DiT component with full-parameter SFT.

\noindent \textbf{Datasets.}
DAVIS\cite{perazzi2016benchmark} and YouTube-VOS\cite{xu2018youtube} are widely used for video inpainting with stationary random masks, but they do not provide ground-truth videos after object removal. Therefore, they are less suitable for VOR evaluation unless synthetic or subjective protocols are introduced. Based on this, we evaluate our model on three datasets: RORD \cite{sagong2022rord}, ROVI \cite{wu2024towards}, and VPLM \cite{yoon2024raccoon}. RORD is a large-scale real-world dataset specifically designed for object removal, containing 2,761 distinct scenes for training and 343 scenes for testing. ROVI includes 5,650 videos with a total of 9,091 object removal instances, among which 458 videos (758 objects) are reserved for testing. From the training splits of RORD and ROVI, we extract 25-frame video segments to form our training set, resulting in approximately 30k samples. For evaluation, we adopt the VPLM dataset, which provides 60 video-based object removal tasks. All three datasets provide triplets of original video, object mask, and ground-truth removed video. 

\noindent \textbf{Metrics.}
To quantitatively evaluate the performance of video object removal, we employ four established metrics: PSNR \cite{hore2010image}, SSIM \cite{wang2004image}, VFID \cite{wang2018video}, and LPIPS \cite{zhang2018unreasonable}. PSNR and SSIM are widely used for measuring low-level pixel-wise similarity between the generated video and the ground truth. To further assess perceptual quality, we use LPIPS, which leverages deep features to better align with human perception. We use VFID, an adaptation of FID for videos, to evaluate the overall temporal consistency and visual realism of the completed video sequences. Additionally, we also introduced flow warping error\cite{lai2018learning} as a metric for measuring temporal consistency in the zero-shot validation. We strictly follow the settings of ProPainter \cite{zhou2023propainter} for calculating metrics.

% 与先进方法的对比（可视化）
% 与扩散模型的方法的时间对比
% 在大掩码区域下的表现 不精细掩码
% 

\begin{table}[t]
\centering
\small
\setlength{\tabcolsep}{2.5pt}
\caption{Quantitative comparisons on RORD, ROVI, and VPLM datasets. $\uparrow$ indicates higher is better. $\downarrow$ indicates lower is better. The best and second performances are highlighted in bold and underlined.}
\label{tab:quantitative_comparisons}
\resizebox{\textwidth}{!}{
\begin{tabular}{lcccccccccccc}
\toprule
\multirow{2}{*}{Models} & \multicolumn{4}{c}{RORD} & \multicolumn{4}{c}{ROVI} & \multicolumn{4}{c}{VPLM} \\
\cmidrule(lr){2-5} \cmidrule(lr){6-9} \cmidrule(l){10-13}
 & PSNR $\uparrow$ & SSIM $\uparrow$ & VFID $\downarrow$ & LPIPS $\downarrow$ & PSNR $\uparrow$ & SSIM $\uparrow$ & VFID $\downarrow$ & LPIPS $\downarrow$ & PSNR $\uparrow$ & SSIM $\uparrow$ & VFID $\downarrow$ & LPIPS $\downarrow$ \\
\midrule
FLoED~\cite{gu2024coherent} & 26.06 & 0.8263 & 0.471 & 0.1557 & 34.42 & 0.9225 & 0.138 & 0.0763 & 34.85 & 0.9149 & 0.493 & 0.0816 \\
E$^2$FGVI~\cite{li2022towards}& 26.95 & 0.8421 & 0.311 & 0.1340 & 38.01 & 0.9693 & 0.096 & 0.0337 & 38.29 & 0.9652 & 0.424 & 0.0390 \\
ROSE~\cite{miao2025rose} & 27.94 & 0.8959 & 0.270 & 0.1105 & 35.72 & 0.9642 & 0.099 &  0.0379 & 36.19 & 0.9520 & 0.378 & 0.0473 \\
FuseFormer~\cite{liu2021fuseformer}  & 27.99 & 0.8659 & 0.306 & 0.1431 & 41.21 & 0.9813 & 0.064 & 0.0292 & 41.24 & 0.9814 & 0.236 & 0.0364 \\
DiffuEraser~\cite{li2025diffueraser} & 28.64 & 0.8865 & 0.268 & 0.1205 & 38.79 & 0.9769 & 0.093 & 0.0323 & 38.76 & 0.9719 & 0.298 & 0.0421\\
FGT~\cite{zhang2022flow} & 30.13 & 0.8972 & 0.271 & 0.1042 & 37.80 & 0.9725 & 0.089 & 0.0325 & 38.51 & 0.9725 & 0.350 & 0.0368 \\
ProPainter~\cite{zhou2023propainter} & 30.97 & 0.9132 & \underline{0.230} & 0.1020 & 40.86 & 0.9759 & 0.069 &\underline{0.0263}& 41.15 & 0.9808 & 0.241 & 0.0374  \\
MiniMax-Remover~\cite{zi2025minimax} & 30.60 & 0.9166 & \textbf{0.224} & \underline{0.0947} & 38.65 & 0.9775 & 0.069 & 0.0318 & 39.47 & 0.9744 & 0.246 & 0.0381 \\
\midrule
\textbf{D2DF-DG(Ours)} & \textbf{32.20} & \textbf{0.9318} & 0.251 & \textbf{0.0902} & \textbf{42.41} & \underline{0.9888} & \textbf{0.062} & \textbf{0.0227} & \textbf{43.31} & \textbf{0.9877} & \textbf{0.191} & \textbf{0.0337} \\
\textbf{D2DF-DF(Ours)} & \underline{31.56} & \underline{0.9202} & 0.268 & 0.1001 & \underline{42.25} & \textbf{0.9889} & \underline{0.063} & 0.0282 & \underline{43.05} & \underline{0.9864} & \underline{0.196} & \underline{0.0348} \\
\bottomrule
\end{tabular}
}
\label{tab:compare}
\end{table}

\noindent
\begin{minipage}{0.5\textwidth}
  \centering
  % 这里改用普通的 minipage 分左右，避开 subfigure 报错
  \begin{minipage}{0.48\linewidth}
    \centering
    \includegraphics[width=\linewidth]{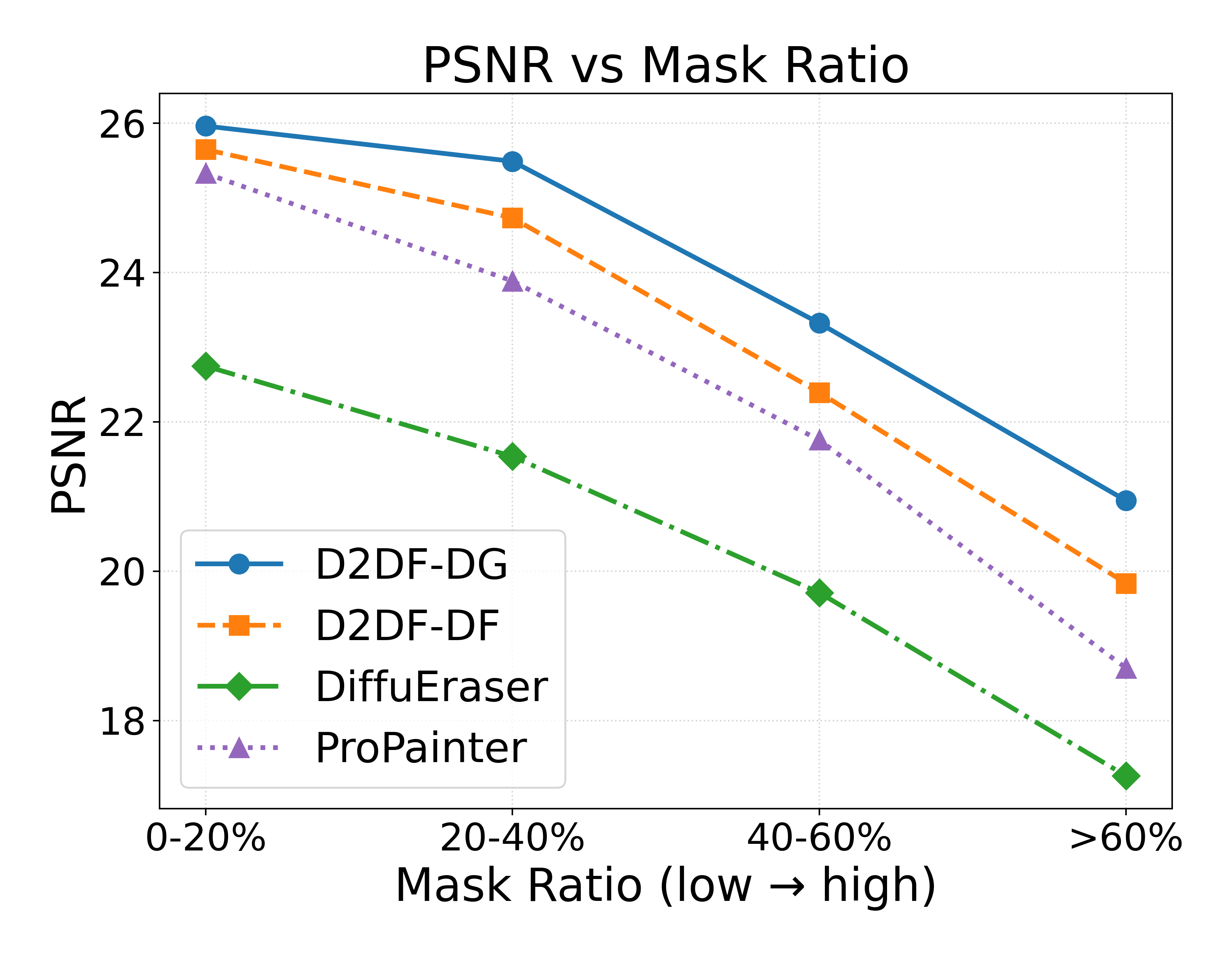}
    \centerline{\footnotesize (a) PSNR}
  \end{minipage}
  \hfill
  \begin{minipage}{0.48\linewidth}
    \centering
    \includegraphics[width=\linewidth]{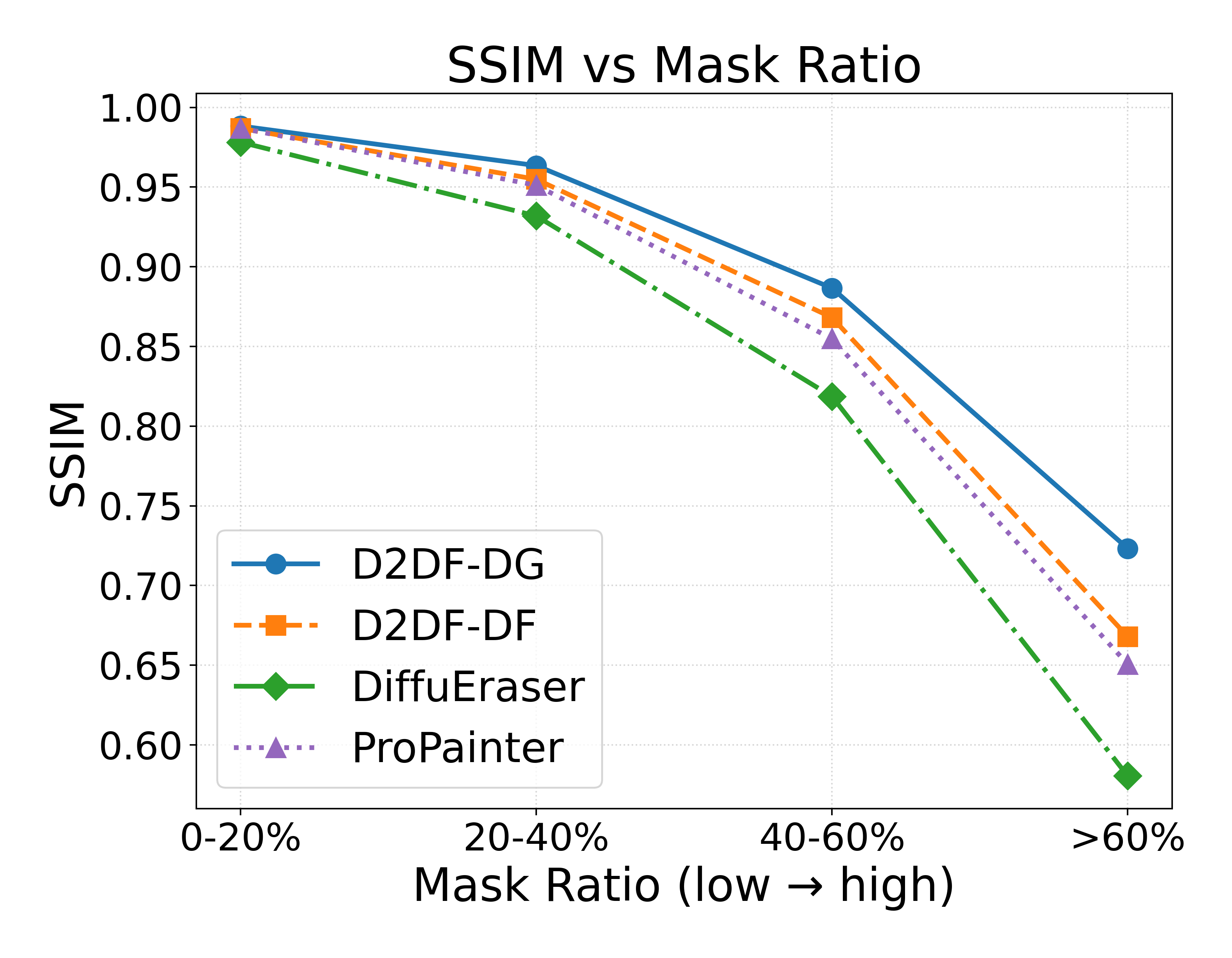}
    \centerline{\footnotesize (b) SSIM}
  \end{minipage}
  \captionof{figure}{Comparison of different models at different mask ratios.}
  \label{fig:compare_psnr_ssim}
\end{minipage}
\hfill
\begin{minipage}{0.48\textwidth}
\setlength{\tabcolsep}{3pt}
\centering
\small
\captionof{table}{Efficiency comparison of diffusion-based methods. ``Prior'' indicates the inference time of the prior extraction stage. ``Denoising'' indicates the denoising time. ``Step'' indicates the number of iterations for denoising.}
\label{tab:speed_comparison}
\resizebox{\textwidth}{!}{
\begin{tabular}{lcccc}
\toprule
Models & Prior(s) &  Denoise(s) & Steps & Total Time(s) \\
\midrule
ROSE\cite{miao2025rose} & -- & 45.76 & 50 & 45.76 \\
FLoED\cite{gu2024coherent} & 2.66 & 31.49 & 25 & 34.15 \\
DiffuEraser\cite{li2025diffueraser} & 2.38 & 2.85 & 2 & 5.13 \\
\textbf{D2DF-DG} & 2.38 & 1.03 & \textbf{1} & 3.41 \\
\textbf{D2DF-DF}& \textbf{0.02} & \textbf{1.03} & \textbf{1} & \textbf{1.05} \\
\bottomrule
\end{tabular}
}
\end{minipage}

\subsection{Comparisons}
\noindent \textbf{Quantitative Evaluation.} We report quantitative results on three datasets. We compare the proposed D2DF-DG and D2DF-DF with previous video object removal methods, including FuseFormer \cite{liu2021fuseformer}, FGT \cite{zhang2022flow}, E$^2$FGVI \cite{li2022towards}, FLoED \cite{gu2024coherent}, DiffuEraser\cite{li2025diffueraser}, ROSE \cite{miao2025rose} and ProPainter \cite{zhou2023propainter}. Among them, FLoED, DiffuEraser, and ROSE are methods based on diffusion models. As shown in Table.\ref{tab:compare}, D2DF-DG achieves the best performance on most metrics and consistently delivers strong results across all three benchmarks. Although D2DF-DF exhibits slightly lower performance compared to D2DF-DG, it still leads state-of-the-art methods in PSNR. This fully demonstrates the powerful capability of our approach in object removal. Additionally, we observe a significant number of large objects in the RORD dataset. Therefore, we select the best models from diffusion-based (DiffuEraser) and traditional approaches (Propainter) and compare their performance at different mask ratios. As shown in Fig.\ref{fig:compare_psnr_ssim}, our models demonstrate only a marginal advantage over Propainter and DiffuEraser within the 0-20\% mask ratio range. However, as the mask ratio increases, our models show a more pronounced advantage in both PSNR and SSIM metrics. This highlights the robustness of our model and its superiority in removing large-area objects.

\begin{figure*}[!t]
    \centering
    \includegraphics[width=\linewidth]{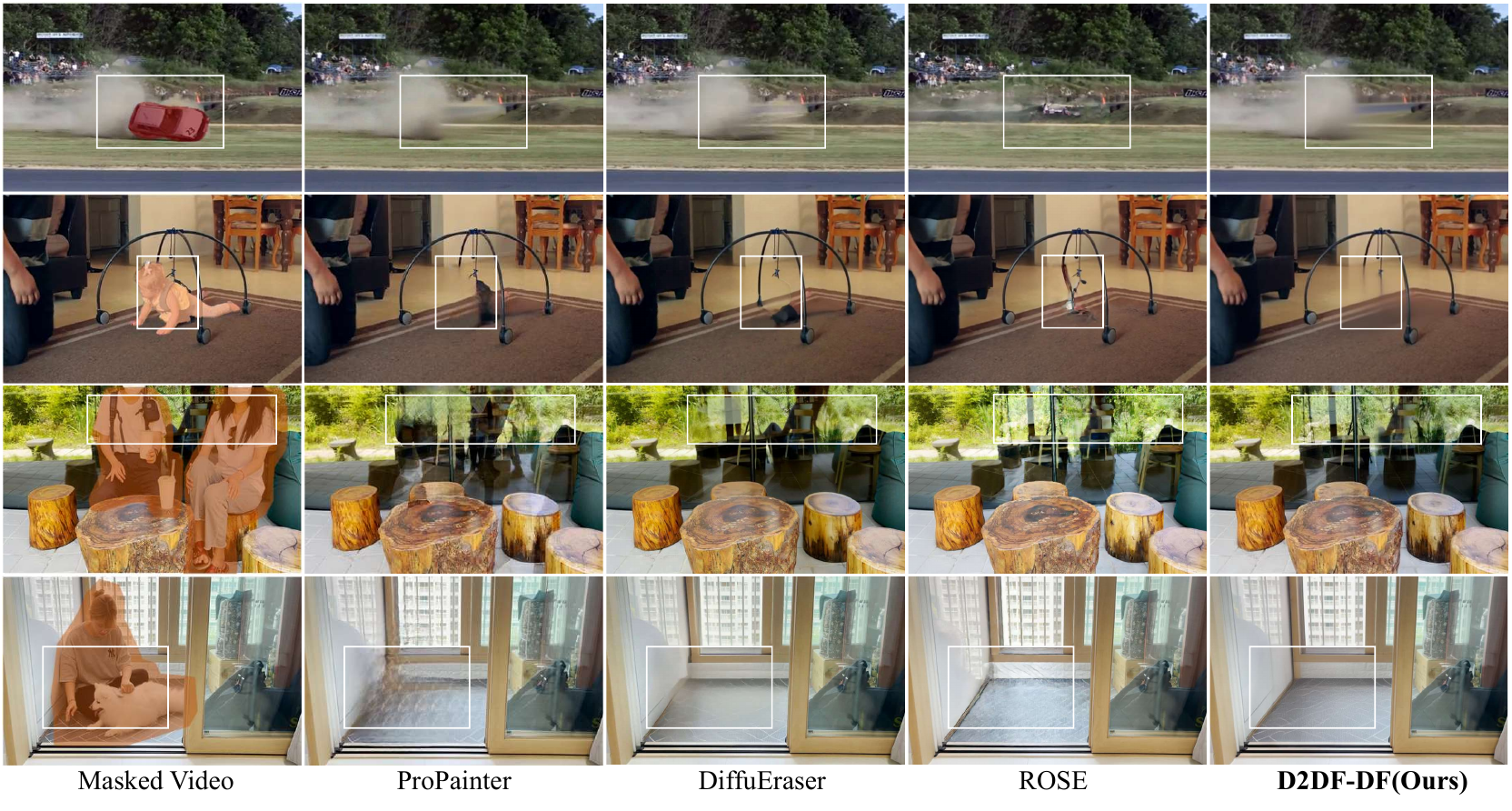}
    \caption{Qualitative comparisons on video object removal. Our D2DF-DF model demonstrates exceptionally strong object removal and background reconstruction capabilities. }
    % \vspace{-0.5cm}
    \label{fig:qualitative_compare}
\end{figure*}

\noindent \textbf{Qualitative Evaluation.} 
For qualitative evaluation, we select diffusion-based methods ROSE \cite{miao2025rose} and DiffuEraser \cite{li2025diffueraser} alongside the lightweight transformer-based ProPainter \cite{zhou2023propainter} for comparison. As shown in Fig.\ref{fig:qualitative_compare}, our method D2DF-DF demonstrates strong capabilities for reconstructing background regions. When removing objects from larger areas in the last two rows, D2DF-DF still delivers highly stable performance. However, other methods exhibit issues such as inconsistent texture information, blurring, and artifacts. This highlights the powerful performance of our one-step denoising model in video object removal. The D2DF-DG delivers even more powerful results. Due to space limitations, you can refer to the appendix for details.

\noindent \textbf{Efficiency Comparison with Diffusion-based Methods.}
Tab.\ref{tab:speed_comparison} compares the inference times of different diffusion-based methods. We test each model on a NVIDIA A100 GPU to generate 25 frames of 480$\times$720 video, measuring the time required for both prior extraction and denoising. Results indicate that DiffuEraser \cite{li2025diffueraser}, also utilizing ProPainter, achieves performance extremely close to our D2DF-DG. However, when switching to our SGFP module for prior extraction, we observed that this module completes the task in just 0.02 seconds. Consequently, D2DF-DF delivers the fastest inference time while maintaining superior performance. Including the time for encoding and decoding latents, our complete processing time is within 10 seconds.

\begin{figure*}[h]
    \centering
    \includegraphics[width=\linewidth]{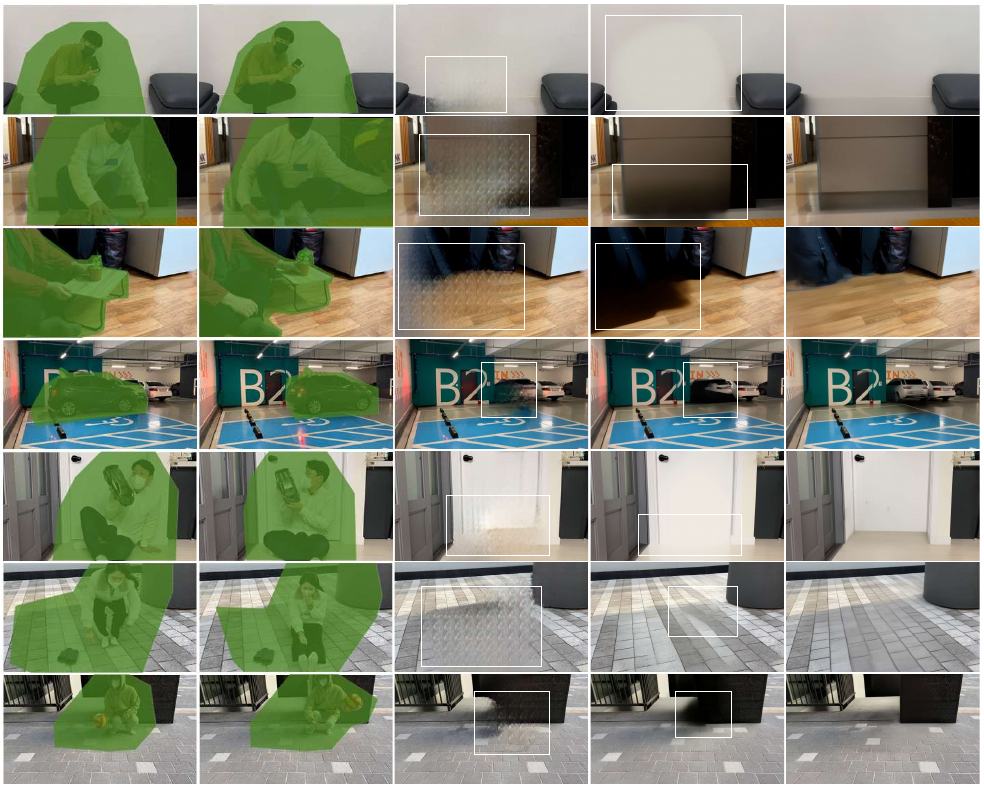}
    \begin{minipage}[t]{0.19\linewidth}
            \centering (a) First Frame
        \end{minipage}
        \hfill
        \begin{minipage}[t]{0.18\linewidth} %稍微调整宽度分配以适应内容
            \centering (b) Last Frame
        \end{minipage}
        \hfill
        \begin{minipage}[t]{0.16\linewidth}
            \centering (c) ProPainter
        \end{minipage}
        \hfill
        \begin{minipage}[t]{0.24\linewidth}
            \centering (d) MiniMax-Remover
        \end{minipage}
        \begin{minipage}[t]{0.19\linewidth}
            \centering (e) D2DF-DF
        \end{minipage}
    \caption{Qualitative comparisons under extreme occlusion conditions. Our DF model achieves remarkable structural consistency and generative performance in one step. }
    \vspace{-0.5cm}
    \label{fig:extreme_case}
\end{figure*}

\noindent \textbf{Qualitative Analysis Under Extreme Occlusion Conditions.} In this section, we will discuss how different models perform under extreme occlusion conditions. When objects in a video have a limited range of motion or occupy a large area, the background regions occluded by these objects cannot be obtained through frame propagation. We define this scenario as extreme occlusion. RORD \cite{sagong2022rord} employs a polygonal mask annotation method, resulting in a dataset with an exceptionally high number of extreme occlusions, making it highly challenging. We selected the state-of-the-art methods ProPainter\cite{zhou2023propainter} and MiniMax-Remover\cite{zi2025minimax}, based on optical flow and diffusion models for comparison. As shown in Fig.\ref{fig:extreme_case}, for regions where information cannot propagate between frames, ProPainter directly produces blurred textures, while MiniMax-Remover (6 steps) exhibits noise artifacts. In contrast, our D2DF-DF model achieves excellent background reproduction with just one step. This demonstrates a unique advantage of our model under extreme occlusion conditions.

\begin{table}[t]
\small
\centering
\caption{Zero-shot performance on Camera-Bench\cite{liu2026understanding}. ``Draft'' denotes whether the method requires an external draft during inference, and ``Flow'' denotes whether optical flow is involved in its inference pipeline. $E_{\mathrm{warp}}$ is reported in $10^{-4}$. The best results are highlighted in bold.}
\label{tab:camera_bench}
\resizebox{0.70\linewidth}{!}{
\begin{tabular}{lcc|cccc}
\toprule
Method & Draft & Flow & PSNR$\uparrow$ & SSIM$\uparrow$ & LPIPS$\downarrow$ & $E_{\mathrm{warp}}\downarrow(\times10^{-4})$ \\
\midrule
FuseFormer~\cite{liu2021fuseformer} & -- & \XSolidBrush  & 25.05 & 0.9155 & 0.2020 & 0.70 \\
ProPainter~\cite{zhou2023propainter} & -- & \Checkmark & 25.42 & 0.9203 & 0.1268 & 1.10 \\
ROSE~\cite{miao2025rose} & -- & \XSolidBrush  & 26.45 & 0.9286 & 0.0860 & 0.55 \\
MiniMax-Remover~\cite{zi2025minimax} & -- & \XSolidBrush  & 26.63 & 0.9324 & 0.0854 & 0.37 \\
\hline
\textbf{D2DF-DG(Ours)} & FuseFormer & \XSolidBrush  & \textbf{27.08} & 0.9421 & \textbf{0.0831} & 0.41 \\
\textbf{D2DF-DG(Ours)} & ProPainter & \Checkmark & 26.96 & \textbf{0.9429} & 0.0884 & 0.93 \\
\textbf{D2DF-DF(Ours)} & -- & \XSolidBrush & 26.57 & 0.9365 & 0.0896 & \textbf{0.27} \\
\bottomrule
\end{tabular}
}
\end{table}

\begin{figure*}[t]
  \centering
  \scriptsize % 建议缩小字号以防换行
    \begin{minipage}[t]{0.24\linewidth}
        \centering  Original Video
    \end{minipage}
    \begin{minipage}[t]{0.24\linewidth} %稍微调整宽度分配以适应内容
        \centering  DiffuEraser
    \end{minipage}
    \begin{minipage}[t]{0.245\linewidth}
        \centering ProPainter
    \end{minipage}
    \begin{minipage}[t]{0.245\linewidth}
        \centering FuseFormer
    \end{minipage}
  \begin{subfigure}{0.245\linewidth}
    \includegraphics[width=\linewidth]{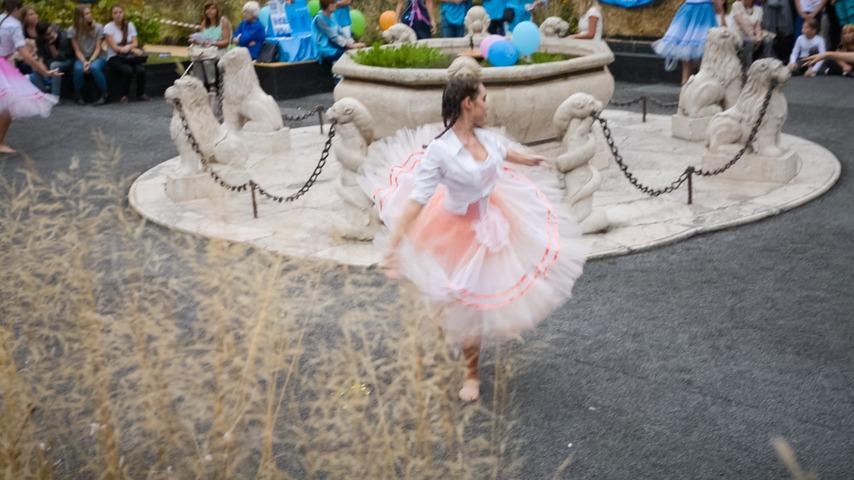}
  \end{subfigure}
  \hspace{-3pt}
  \begin{subfigure}{0.245\linewidth}
    \includegraphics[width=\linewidth]{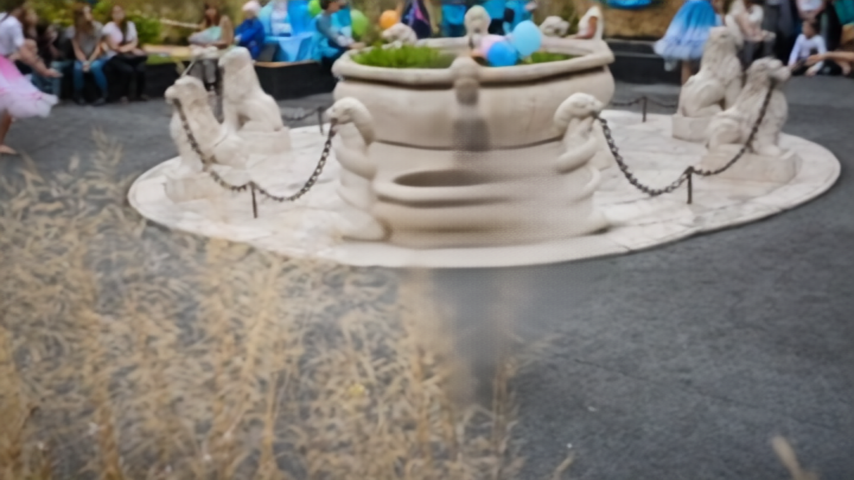}
  \end{subfigure}
  \hspace{-3pt}
  \begin{subfigure}{0.245\linewidth}
    \includegraphics[width=\linewidth]{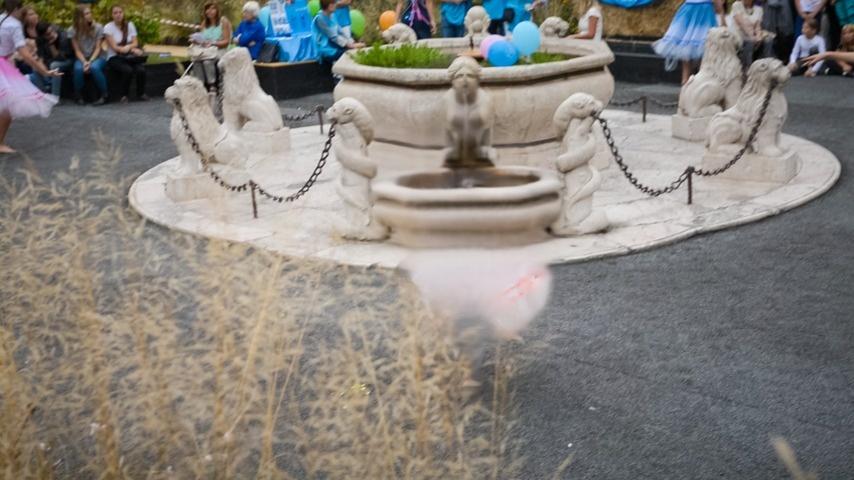}
  \end{subfigure}
  \hspace{-3pt}
  \begin{subfigure}{0.245\linewidth}
    \includegraphics[width=\linewidth]{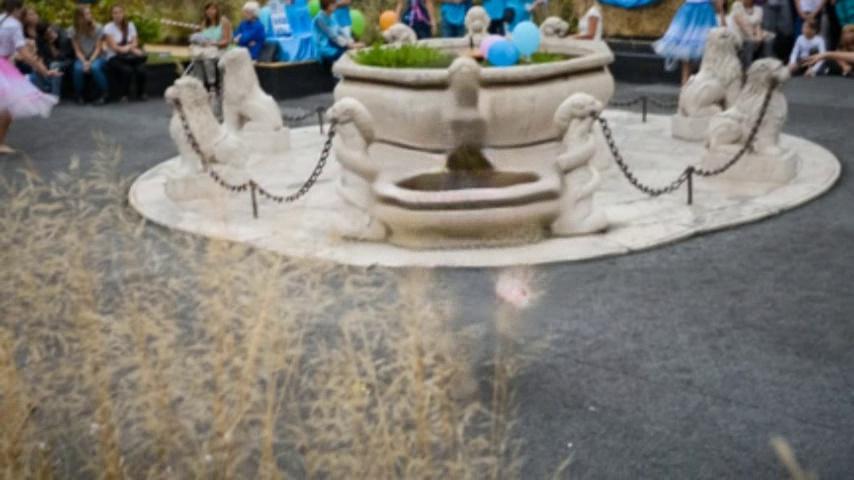}
  \end{subfigure}
  \begin{subfigure}{0.245\linewidth}
    \includegraphics[width=\linewidth]{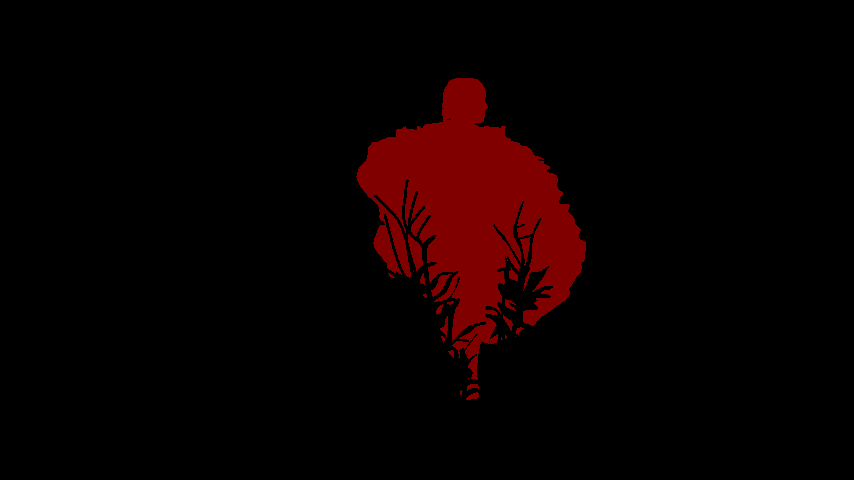}
  \end{subfigure}
  \hspace{-3pt}
  \begin{subfigure}{0.245\linewidth}
    \includegraphics[width=\linewidth]{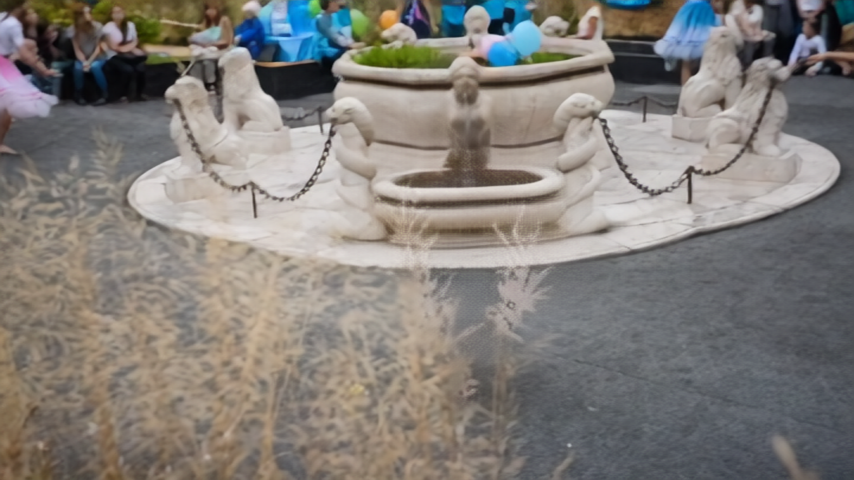}
  \end{subfigure}
  \hspace{-3pt}
  \begin{subfigure}{0.245\linewidth}
    \includegraphics[width=\linewidth]{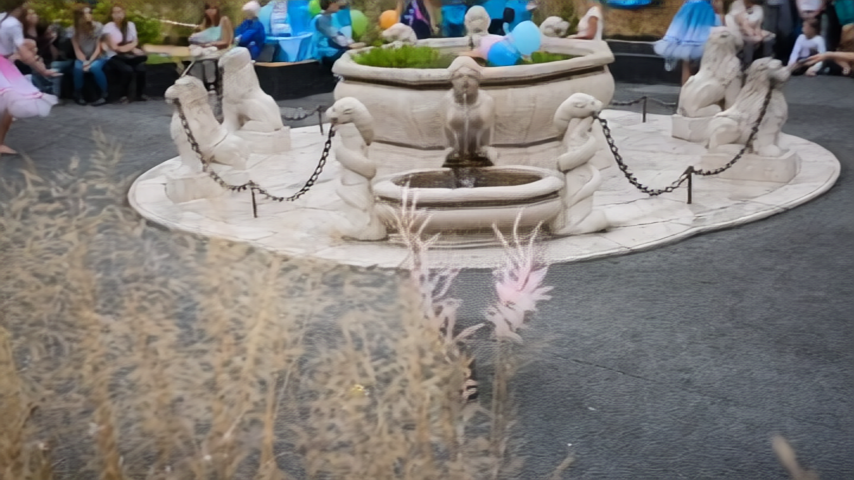}
  \end{subfigure}
  \hspace{-3pt}
  \begin{subfigure}{0.245\linewidth}
    \includegraphics[width=\linewidth]{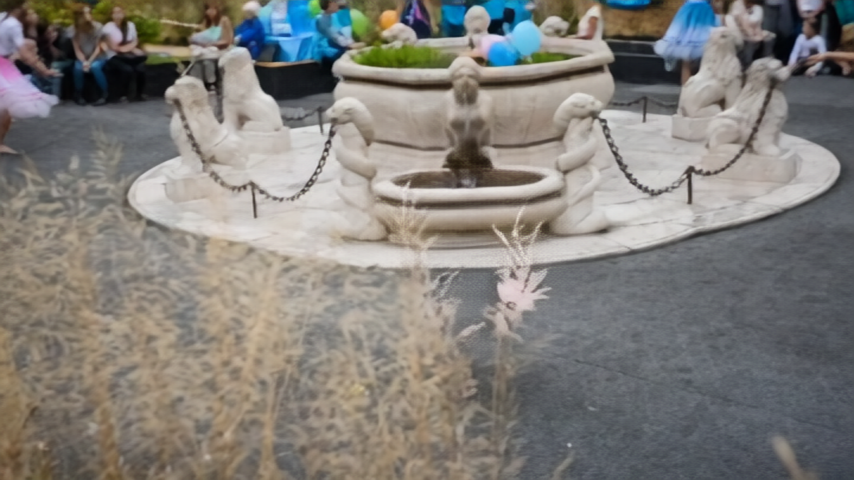}
  \end{subfigure}

  \scriptsize % 建议缩小字号以防换行
    \begin{minipage}[t]{0.24\linewidth}
        \centering  Mask
    \end{minipage}
    \begin{minipage}[t]{0.24\linewidth} %稍微调整宽度分配以适应内容
        \centering  DG(DiffuEraser)
    \end{minipage}
    \begin{minipage}[t]{0.245\linewidth}
        \centering DG(ProPainter)
    \end{minipage}
    \begin{minipage}[t]{0.245\linewidth}
        \centering DG(FuseFormer)
    \end{minipage}
  \caption{Visual verification of the generalization of draft sensitivity on a sample from DAVIS. The results demonstrate that our DG can refine various types of drafts.}
  \label{fig:draft_source}
  \vspace{-10pt}
\end{figure*}

\subsection{Robustness and Generalization Analysis}
\noindent\textbf{Zero-shot Evaluation on Camera-Bench.}
To further evaluate the generalization ability of our method, we conduct zero-shot experiments on Camera-Bench\cite{liu2026understanding}, as shown in Tab.~\ref{tab:camera_bench}. Without training on this benchmark, both D2DF-DG and D2DF-DF remain highly competitive against existing methods. It is worth noting that our framework is not inherently tied to optical flow. Whether optical flow is involved depends on the draft source used by D2DF-DG. Although D2DF-DG is trained with ProPainter drafts, it can still effectively refine FuseFormer drafts, which are generated without optical flow. This suggests that the cross-domain robustness of our method is not limited to dataset transfer, but also extends to different types of draft priors. Furthermore, D2DF-DF removes the external draft dependency entirely, making the final model fully free from optical-flow-related concerns during inference.

\noindent\textbf{Draft-source Robustness.}
We further evaluate D2DF-DG with drafts from different sources. As shown in Fig.~\ref{fig:draft_source}, although our model is trained with ProPainter drafts, it can consistently refine drafts generated by DiffuEraser, ProPainter, and FuseFormer, producing similarly high-quality removal results. This demonstrates that D2DF-DG does not overfit to a specific draft distribution, but learns a robust refinement ability across heterogeneous draft priors. Notably, the effectiveness on FuseFormer drafts also indicates that our framework can work with flow-free draft sources, while D2DF-DF further removes the draft dependency entirely.

\subsection{Ablation Study}

\noindent \textbf{Study of Prior-Privileged Consistency Distillation.}
In Tab.\ref{tab:ablation_ppcd}, when we train directly using the LDM approach, the model behaves as a standard LDM model, yielding suboptimal results in a single step. Consequently, employing Consistency Distillation (CD) significantly improves performance. However, introducing privileged prior knowledge (replacing the draft conditions with ground truth) further enhances results. This indicates that our ground-truth prior indeed yields a more stable consistency trajectory.

\begin{table}[t]
\centering
\setlength\tabcolsep{5pt}
\small
\vspace{-5pt}
\caption{Ablation study of proposed Prior-Privileged Consistency Distillation (PPCD). “D.T." means directly train the diffusion model as Eq.\ref{eq:ddpm_loss}, “CD" means the Consistency Distillation \cite{song2023consistency}. }
\label{tab:ablation_ppcd}
\begin{tabular*}{0.70 \textwidth}{@{\extracolsep{\fill}}ccccccc@{}}
\toprule
\multicolumn{3}{c}{Training Methods} & 
\multicolumn{4}{c}{Metrics} \\
\cmidrule(lr){1-3} \cmidrule(lr){4-7}
D.T. & CD & PPCD & PSNR$\uparrow$  & SSIM$\uparrow$  & VFID$\downarrow$  & LPIPS$\downarrow$   \\
\midrule
\checkmark &  &  & 31.54 & 0.9192 & 0.271 &   0.1014  \\
 & \checkmark &  & 31.85 & 0.9296 & 0.257 &  0.1002  \\
 &  & \checkmark & \textbf{32.20} & \textbf{0.9318} & \textbf{0.251} &  \textbf{0.0902}  \\
\bottomrule
\end{tabular*}
\end{table}

\begin{table}[t]
\centering
\vspace{-5pt}
\caption{Ablation study of our SGFP module. The first row means the baseline model without any extra prior. “TL" means the transformer layers, ``VP'' indicates whether to exclude masked patches as valid patches, ``SW'' indicates the spatial window constraint, and ``MR'' indicates reconstruction of the masked region only.}
\label{tab:ablation_SGFP}
\small
\begin{tabular*}{0.70 \textwidth}{@{\extracolsep{\fill}}llccccc@{}}
\toprule
\multicolumn{4}{c}{Components} & 
\multicolumn{3}{c}{Metrics} \\
\cmidrule(lr){1-4} \cmidrule(lr){5-7}
TL &  VP & SW & MR & PSNR$\uparrow$  & SSIM$\uparrow$  & LPIPS$\downarrow$   \\
\midrule
& &  & & 29.36 & 0.8920 &   0.1426  \\
\checkmark &  &  &  & 30.73 & 0.9171 &  0.1149  \\
\checkmark & \checkmark &  &  & 30.78 & \textbf{0.9228} &  0.1093 \\
\checkmark & \checkmark &  \checkmark & & 31.47 & 0.9197 & 0.1070 \\
\checkmark & \checkmark & \checkmark & \checkmark & \textbf{31.56} &  0.9202 &  \textbf{0.1001}  \\
\bottomrule
\end{tabular*}
\vspace{-15pt}
\end{table}

\begin{figure}[t]
    \centering
    % --- 左边的图 (Figure 1) ---
    \begin{minipage}[t]{0.54\linewidth}
        \vspace{0pt}
        \centering
        % 图片宽度设为当前minipage的宽度
        \includegraphics[width=\linewidth]{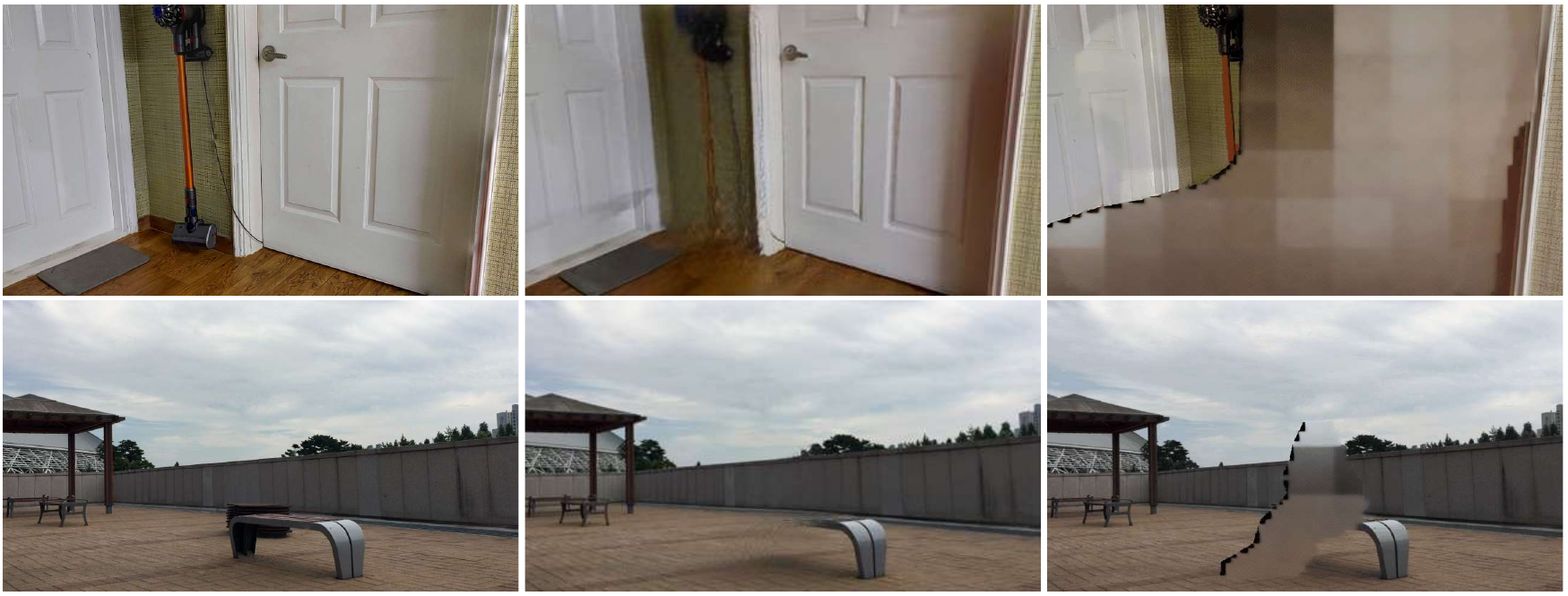}
        
        % 原有的标签布局 (注意字体可能需要调小，因为空间变窄了)
        \scriptsize % 建议缩小字号以防换行
        \begin{minipage}[t]{0.32\linewidth}
            \centering (a) Ground Truth
        \end{minipage}
        \hfill
        \begin{minipage}[t]{0.28\linewidth} %稍微调整宽度分配以适应内容
            \centering (b) Draft
        \end{minipage}
        \hfill
        \begin{minipage}[t]{0.36\linewidth}
            \centering (c) Pseudo-Draft
        \end{minipage}
        
        \caption{Visualization of Drafts and Pseudo-Drafts. (a) Ground Truth. (b) Drafts with artifacts. (c) Pseudo-drafts from decoder.}
        \label{fig:draft_pseudo}
    \end{minipage}
    \hfill % 左右两个minipage之间的弹簧空格
    % --- 右边的图 (Figure 2) ---
    \begin{minipage}[t]{0.44\linewidth}
        \vspace{0pt}
        \centering
        \includegraphics[width=\linewidth]{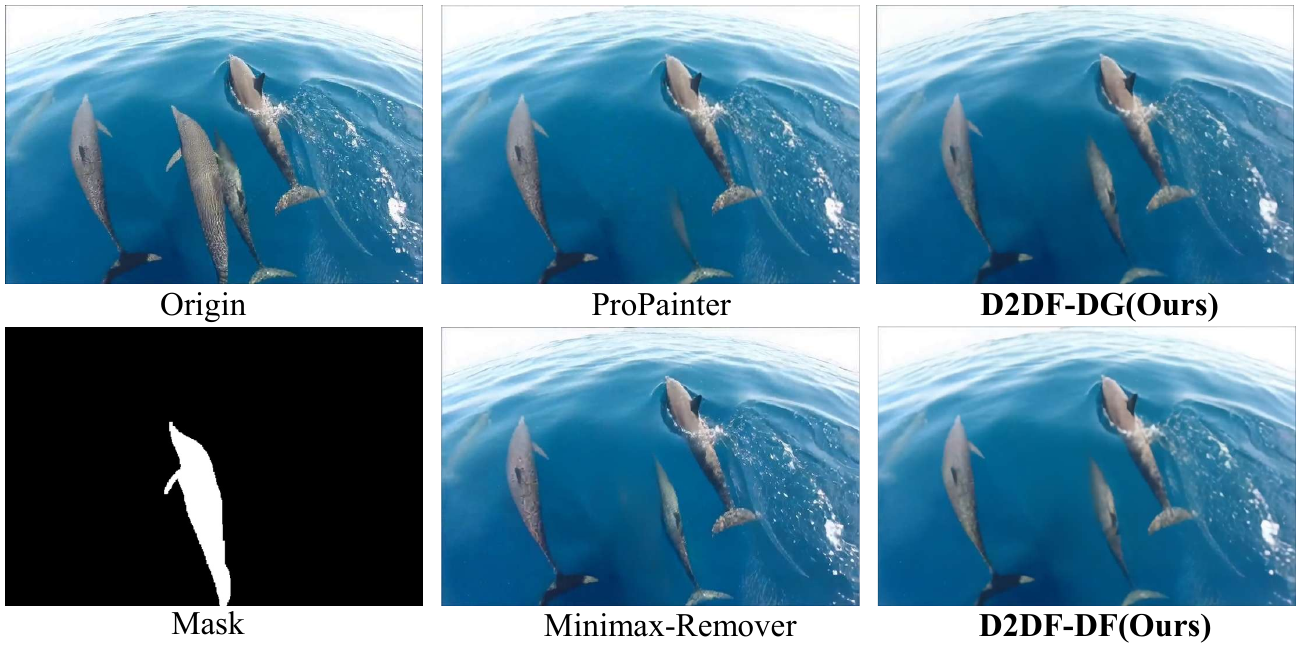}
        \caption{Analysis of failure cases. Object removal models fail when complex occlusions exist between objects.}
        \label{fig:failurecase}
    \end{minipage}
\end{figure}

\noindent \textbf{Study of Self-Guided Fast Planting Module.}
To investigate the effectiveness of our proposed SGFP module, we conduct detailed ablation experiments on the RORD dataset. As shown in Tab.\ref{tab:ablation_SGFP}, the first row represents the baseline model, indicating the performance without any prior knowledge. Adding the transformer layer (TL) yields an initial improvement in model performance. Excluding masked regions (VP) results in a slight further enhancement. Introducing the spatial window (SW) significantly boosts model effectiveness. Finally, reconstructing only the masked region (MR) achieves optimal model performance. These ablation experiments demonstrate the rationality and effectiveness of each design step. To validate that the latents generated by SGFP indeed contain meaningful background information, we additionally trained a diagnostic decoder. This decoder is solely used for visualization. As shown in Fig.\ref{fig:draft_pseudo}(c), our pseudo-drafts exhibit blocky filling because they are learned from many patch latents. Despite its visual coarseness compared to the draft in Fig.\ref{fig:draft_pseudo}(b), this decoding result demonstrates that SGFP successfully captures key background color tones and structures, which suffice as effective prior information.

\noindent \textbf{Study of Three-stage Distillation Framework. }
To validate the effectiveness of the three-stage framework, we conducted experiments as shown in Table.\ref{tab:ablation_3stage}. As the table indicates, the performance achieved when the teacher network is LDM was not as good as that obtained with D2DF-DG as the teacher network. This shows the trained D2DF-DG model can provide more stable and consistent solution trajectories than the diffusion D-LDM. Furthermore, the comparison between CD and PPCD also demonstrated the superiority of the PPCD method.

% \begin{figure}[t]
%     \centering
%     \includegraphics[width=\linewidth]{figures/visual_draft.pdf}
%     \begin{minipage}[t]{0.33\linewidth}
%     \centering
%     (a) Ground Truth
%   \end{minipage}
%   \hfill
%   \begin{minipage}[t]{0.2\linewidth}
%     \centering
% (b) Draft
%   \end{minipage}
%   \hfill
%   \begin{minipage}[t]{0.35\linewidth}
%     \centering
% (c) Pseudo-Draft
%   \end{minipage}
%     \caption{Visualization of Drafts and Pseudo-Drafts. (a) Ground Truth of the removed video frames. (b) Drafts with artifacts. (c) Pseudo-drafts decoded by the diagnostic decoder.}
%     % \vspace{-0.5cm}
%     \label{fig:draft_pseudo}
% \end{figure}

\begin{table}[t]
\centering
\setlength{\tabcolsep}{5pt}
\small
\caption{Ablation study of our three-stage framework to obtain D2DF-DF. “CD" means the Consistency Distillation \cite{song2023consistency}. The best results are achieved by using PPCD for distilling D2DF-DG. }
\label{tab:ablation_3stage}
\begin{tabular*}{0.75 \textwidth}{@{\extracolsep{\fill}}llccccc@{}}
\toprule
\multicolumn{4}{c}{Training Settings} & 
\multicolumn{3}{c}{Metrics} \\
\cmidrule(lr){1-4} \cmidrule(lr){5-7}
Student & Teacher & CD & PPCD & PSNR$\uparrow$  & SSIM$\uparrow$  & LPIPS$\downarrow$   \\
\midrule
\multirow{4}{*}{D2DF-DF}& D-LDM & \checkmark & & 31.3268 & 0.9179 &  0.1025  \\
& D-LDM &  & \checkmark & 31.4484 & 0.9197 &  0.1014  \\
& D2DF-DG &  \checkmark &  & 31.3257 & 0.9173 & 0.1036  \\
& D2DF-DG &  & \checkmark & \textbf{31.5646} & \textbf{0.9202} & \textbf{0.1001}   \\
\bottomrule
\end{tabular*}
\vspace{-10pt}
\end{table}

\subsection{Analysis of Failure Cases}

In this section, we will discuss cases where the model fails to remove objects and analyze the reasons for these failures. We observe that when multiple objects exist in the video and these objects exhibit dynamic and complex occlusion relationships, the model is unable to fully reconstruct the occluded objects. As shown in Fig.\ref{fig:failurecase}, we compared the state-of-the-art optical flow-based method ProPainter\cite{zhou2023propainter} and the state-of-the-art diffusion model-based method MiniMax-Remover\cite{zi2025minimax}. The results indicate that neither method can fully restore the occluded ``dolphin'' in this example. Our two models, D2DF-DG and D2DF-DF, also produce imperfect results under one-step inference settings. However, they retain significantly more context compared to MiniMax-Remover. This also demonstrates the superiority of our approach.

\vspace{-10pt}
\section{Conclusion}
\vspace{-5pt}
In this paper, we present D2DF, a unified framework that progressively evolves from a draft-guided diffusion teacher to a fully draft-free one-step video object removal model. By introducing Prior-Privileged Consistency Distillation (PPCD), we effectively stabilize the distillation process and enable high-quality one-step generation. Furthermore, the proposed Self-Guided Fast Planting (SGFP) module eliminates the dependency on external draft inputs by generating scene-consistent pseudo-drafts directly in latent space. Our method not only excels in visual quality and temporal coherence but also achieves remarkable efficiency. However, the blurring that occurs in certain scenarios with the one-step denoising is an enhancement needed in future work.

\vspace{-5pt}
\subsubsection*{Acknowledgements.} 
Zizhao Chen and Ping Wei acknowledge the support of the National Natural Science Foundation of China (No. U23B2060, No. 62495092).
Mengmeng Wang acknowledges the support of the National Natural Science Foundation of China (No. 62403429).
We also acknowledge the computational resources provided by SGIT AI Lab, State Grid Corporation of China.

\clearpage  % TODO FINAL: This \clearpage needs to be removed from both review and camera-ready versions.

\bibliographystyle{splncs04}
\bibliography{main}

@String(CVPR  = {IEEE Conf. Comput. Vis. Pattern Recog.})

@String(ECCV  = {Eur. Conf. Comput. Vis.})

@String(BMVC  = {Brit. Mach. Vis. Conf.})

@String(AAAI  = {AAAI})

@String(CVPR  = {CVPR})

@String(ECCV  = {ECCV})

@String(BMVC  =	{BMVC})

@String(CVPR= {IEEE Conf. Comput. Vis. Pattern Recog.})

@String(ECCV= {Eur. Conf. Comput. Vis.})

@String(BMVC= {Brit. Mach. Vis. Conf.})

@String(AAAI = {AAAI})

@article{ebdelli2015video,
  title={Video inpainting with short-term windows: application to object removal and error concealment},
  author={Ebdelli, Mounira and Le Meur, Olivier and Guillemot, Christine},
  journal={IEEE Transactions on Image Processing},
  volume={24},
  number={10},
  pages={3034--3047},
  year={2015},
  publisher={IEEE}
}

@article{wang2023unsupervised,
  title={Unsupervised Temporal Correspondence Learning for Unified Video Object Removal},
  author={Wang, Zhongdao and Wang, Jinglu and Li, Xiao and Li, Ya-Li and Lu, Yan and Wang, Shengjin},
  journal={IEEE Transactions on Image Processing},
  year={2023},
  publisher={IEEE}
}

@inproceedings{zhang2020autoremover,
  title={Autoremover: Automatic object removal for autonomous driving videos},
  author={Zhang, Rong and Li, Wei and Wang, Peng and Guan, Chenye and Fang, Jin and Song, Yuhang and Yu, Jinhui and Chen, Baoquan and Xu, Weiwei and Yang, Ruigang},
  booktitle={Proceedings of the AAAI Conference on Artificial Intelligence},
  volume={34},
  number={07},
  pages={12853--12861},
  year={2020}
}

@article{tang2011video,
  title={Video inpainting on digitized vintage films via maintaining spatiotemporal continuity},
  author={Tang, Nick C and Hsu, Chiou-Ting and Su, Chih-Wen and Shih, Timothy K and Liao, Hong-Yuan Mark},
  journal={IEEE Transactions on Multimedia},
  volume={13},
  number={4},
  pages={602--614},
  year={2011},
  publisher={IEEE}
}

@inproceedings{bertalmio2001navier,
  title={Navier-stokes, fluid dynamics, and image and video inpainting},
  author={Bertalmio, Marcelo and Bertozzi, Andrea L and Sapiro, Guillermo},
  booktitle={Proceedings of the 2001 IEEE Computer Society Conference on Computer Vision and Pattern Recognition. CVPR 2001},
  volume={1},
  pages={I--I},
  year={2001},
  organization={IEEE}
}

@inproceedings{ding2025homogen,
  title={HomoGen: Enhanced Video Inpainting via Homography Propagation and Diffusion},
  author={Ding, Ding and Pan, Yueming and Feng, Ruoyu and Dai, Qi and Qiu, Kai and Bao, Jianmin and Luo, Chong and Chen, Zhenzhong},
  booktitle={Proceedings of the Computer Vision and Pattern Recognition Conference},
  pages={22953--22962},
  year={2025}
}

@inproceedings{zhou2023propainter,
  title={Propainter: Improving propagation and transformer for video inpainting},
  author={Zhou, Shangchen and Li, Chongyi and Chan, Kelvin CK and Loy, Chen Change},
  booktitle={Proceedings of the IEEE/CVF international conference on computer vision},
  pages={10477--10486},
  year={2023}
}

@inproceedings{li2022towards,
  title={Towards an end-to-end framework for flow-guided video inpainting},
  author={Li, Zhen and Lu, Cheng-Ze and Qin, Jianhua and Guo, Chun-Le and Cheng, Ming-Ming},
  booktitle={Proceedings of the IEEE/CVF conference on computer vision and pattern recognition},
  pages={17562--17571},
  year={2022}
}

@inproceedings{zhang2022flow,
  title={Flow-guided transformer for video inpainting},
  author={Zhang, Kaidong and Fu, Jingjing and Liu, Dong},
  booktitle={European conference on computer vision},
  pages={74--90},
  year={2022},
  organization={Springer}
}

@inproceedings{liu2021fuseformer,
  title={Fuseformer: Fusing fine-grained information in transformers for video inpainting},
  author={Liu, Rui and Deng, Hanming and Huang, Yangyi and Shi, Xiaoyu and Lu, Lewei and Sun, Wenxiu and Wang, Xiaogang and Dai, Jifeng and Li, Hongsheng},
  booktitle={Proceedings of the IEEE/CVF international conference on computer vision},
  pages={14040--14049},
  year={2021}
}

@inproceedings{zeng2020learning,
  title={Learning joint spatial-temporal transformations for video inpainting},
  author={Zeng, Yanhong and Fu, Jianlong and Chao, Hongyang},
  booktitle={European conference on computer vision},
  pages={528--543},
  year={2020},
  organization={Springer}
}

@inproceedings{gao2020flow,
  title={Flow-edge guided video completion},
  author={Gao, Chen and Saraf, Ayush and Huang, Jia-Bin and Kopf, Johannes},
  booktitle={European Conference on Computer Vision},
  pages={713--729},
  year={2020},
  organization={Springer}
}

@article{li2025diffueraser,
  title={Diffueraser: A diffusion model for video inpainting},
  author={Li, Xiaowen and Xue, Haolan and Ren, Peiran and Bo, Liefeng},
  journal={arXiv preprint arXiv:2501.10018},
  year={2025}
}

@article{gu2024advanced,
  title={Advanced video inpainting using optical flow-guided efficient diffusion},
  author={Gu, Bohai and Luo, Hao and Guo, Song and Dong, Peiran},
  journal={arXiv e-prints},
  pages={arXiv--2412},
  year={2024}
}

@inproceedings{bian2025videopainter,
  title={Videopainter: Any-length video inpainting and editing with plug-and-play context control},
  author={Bian, Yuxuan and Zhang, Zhaoyang and Ju, Xuan and Cao, Mingdeng and Xie, Liangbin and Shan, Ying and Xu, Qiang},
  booktitle={Proceedings of the Special Interest Group on Computer Graphics and Interactive Techniques Conference Conference Papers},
  pages={1--12},
  year={2025}
}

@article{miao2025rose,
  title={ROSE: Remove Objects with Side Effects in Videos},
  author={Miao, Chenxuan and Feng, Yutong and Zeng, Jianshu and Gao, Zixiang and Liu, Hantang and Yan, Yunfeng and Qi, Donglian and Chen, Xi and Wang, Bin and Zhao, Hengshuang},
  journal={arXiv preprint arXiv:2508.18633},
  year={2025}
}

@inproceedings{lee2025video,
  title={Video diffusion models are strong video inpainter},
  author={Lee, Minhyeok and Cho, Suhwan and Shin, Chajin and Lee, Jungho and Yang, Sunghun and Lee, Sangyoun},
  booktitle={Proceedings of the AAAI Conference on Artificial Intelligence},
  volume={39},
  number={4},
  pages={4526--4533},
  year={2025}
}

@inproceedings{rombach2022high,
  title={High-resolution image synthesis with latent diffusion models},
  author={Rombach, Robin and Blattmann, Andreas and Lorenz, Dominik and Esser, Patrick and Ommer, Bj{\"o}rn},
  booktitle={Proceedings of the IEEE/CVF conference on computer vision and pattern recognition},
  pages={10684--10695},
  year={2022}
}

@article{yang2024cogvideox,
  title={Cogvideox: Text-to-video diffusion models with an expert transformer},
  author={Yang, Zhuoyi and Teng, Jiayan and Zheng, Wendi and Ding, Ming and Huang, Shiyu and Xu, Jiazheng and Yang, Yuanming and Hong, Wenyi and Zhang, Xiaohan and Feng, Guanyu and others},
  journal={arXiv preprint arXiv:2408.06072},
  year={2024}
}

@inproceedings{peebles2023scalable,
  title={Scalable diffusion models with transformers},
  author={Peebles, William and Xie, Saining},
  booktitle={Proceedings of the IEEE/CVF international conference on computer vision},
  pages={4195--4205},
  year={2023}
}

@article{song2020denoising,
  title={Denoising diffusion implicit models},
  author={Song, Jiaming and Meng, Chenlin and Ermon, Stefano},
  journal={arXiv preprint arXiv:2010.02502},
  year={2020}
}

@inproceedings{song2023consistency,
  title={Consistency models},
  author={Song, Yang and Dhariwal, Prafulla and Chen, Mark and Sutskever, Ilya},
  booktitle={Proceedings of the 40th International Conference on Machine Learning},
  pages={32211--32252},
  year={2023}
}

@article{luo2023latent,
  title={Latent consistency models: Synthesizing high-resolution images with few-step inference},
  author={Luo, Simian and Tan, Yiqin and Huang, Longbo and Li, Jian and Zhao, Hang},
  journal={arXiv preprint arXiv:2310.04378},
  year={2023}
}

@article{song2023improved,
  title={Improved techniques for training consistency models},
  author={Song, Yang and Dhariwal, Prafulla},
  journal={arXiv preprint arXiv:2310.14189},
  year={2023}
}

@inproceedings{sagong2022rord,
  title={RORD: A Real-world Object Removal Dataset.},
  author={Sagong, Min-Cheol and Yeo, Yoon-Jae and Jung, Seung-Won and Ko, Sung-Jea},
  booktitle={BMVC},
  pages={542},
  year={2022}
}

@inproceedings{wu2024towards,
  title={Towards language-driven video inpainting via multimodal large language models},
  author={Wu, Jianzong and Li, Xiangtai and Si, Chenyang and Zhou, Shangchen and Yang, Jingkang and Zhang, Jiangning and Li, Yining and Chen, Kai and Tong, Yunhai and Liu, Ziwei and others},
  booktitle={Proceedings of the IEEE/CVF Conference on Computer Vision and Pattern Recognition},
  pages={12501--12511},
  year={2024}
}

@inproceedings{yoon2024raccoon,
    title = "{RACC}oo{N}: Versatile Instructional Video Editing with Auto-Generated Narratives",
    author = "Yoon, Jaehong  and
      Yu, Shoubin  and
      Bansal, Mohit",
    booktitle = "Proceedings of the 2025 Conference on Empirical Methods in Natural Language Processing",
    year = "2025",
    publisher = "Association for Computational Linguistics",
    pages = "27960--27996",
   
}

@inproceedings{hore2010image,
  title={Image quality metrics: PSNR vs. SSIM},
  author={Hore, Alain and Ziou, Djemel},
  booktitle={2010 20th international conference on pattern recognition},
  pages={2366--2369},
  year={2010},
  organization={IEEE}
}

@article{wang2004image,
  title={Image quality assessment: from error visibility to structural similarity},
  author={Wang, Zhou and Bovik, Alan C and Sheikh, Hamid R and Simoncelli, Eero P},
  journal={IEEE transactions on image processing},
  volume={13},
  number={4},
  pages={600--612},
  year={2004},
  publisher={IEEE}
}

@inproceedings{zhang2018unreasonable,
  title={The unreasonable effectiveness of deep features as a perceptual metric},
  author={Zhang, Richard and Isola, Phillip and Efros, Alexei A and Shechtman, Eli and Wang, Oliver},
  booktitle={Proceedings of the IEEE conference on computer vision and pattern recognition},
  pages={586--595},
  year={2018}
}

@article{wang2018video,
  title={Video-to-video synthesis},
  author={Wang, Ting-Chun and Liu, Ming-Yu and Zhu, Jun-Yan and Liu, Guilin and Tao, Andrew and Kautz, Jan and Catanzaro, Bryan},
  journal={arXiv preprint arXiv:1808.06601},
  year={2018}
}

@inproceedings{kim2019deep,
  title={Deep video inpainting},
  author={Kim, Dahun and Woo, Sanghyun and Lee, Joon-Young and Kweon, In So},
  booktitle={Proceedings of the IEEE/CVF conference on computer vision and pattern recognition},
  pages={5792--5801},
  year={2019}
}

@inproceedings{xu2019deep,
  title={Deep flow-guided video inpainting},
  author={Xu, Rui and Li, Xiaoxiao and Zhou, Bolei and Loy, Chen Change},
  booktitle={Proceedings of the IEEE/CVF conference on computer vision and pattern recognition},
  pages={3723--3732},
  year={2019}
}

@inproceedings{lee2019copy,
  title={Copy-and-paste networks for deep video inpainting},
  author={Lee, Sungho and Oh, Seoung Wug and Won, DaeYeun and Kim, Seon Joo},
  booktitle={Proceedings of the IEEE/CVF international conference on computer vision},
  pages={4413--4421},
  year={2019}
}

@inproceedings{wang2019video,
  title={Video inpainting by jointly learning temporal structure and spatial details},
  author={Wang, Chuan and Huang, Haibin and Han, Xiaoguang and Wang, Jue},
  booktitle={Proceedings of the AAAI conference on artificial intelligence},
  volume={33},
  number={01},
  pages={5232--5239},
  year={2019}
}

@inproceedings{zhang2022inertia,
  title={Inertia-guided flow completion and style fusion for video inpainting},
  author={Zhang, Kaidong and Fu, Jingjing and Liu, Dong},
  booktitle={Proceedings of the IEEE/CVF conference on computer vision and pattern recognition},
  pages={5982--5991},
  year={2022}
}

@inproceedings{blattmann2023align,
  title={Align your latents: High-resolution video synthesis with latent diffusion models},
  author={Blattmann, Andreas and Rombach, Robin and Ling, Huan and Dockhorn, Tim and Kim, Seung Wook and Fidler, Sanja and Kreis, Karsten},
  booktitle={Proceedings of the IEEE/CVF conference on computer vision and pattern recognition},
  pages={22563--22575},
  year={2023}
}

@article{dhariwal2021diffusion,
  title={Diffusion models beat gans on image synthesis},
  author={Dhariwal, Prafulla and Nichol, Alexander},
  journal={Advances in neural information processing systems},
  volume={34},
  pages={8780--8794},
  year={2021}
}

@inproceedings{jin2025flovd,
  title={Flovd: Optical flow meets video diffusion model for enhanced camera-controlled video synthesis},
  author={Jin, Wonjoon and Dai, Qi and Luo, Chong and Baek, Seung-Hwan and Cho, Sunghyun},
  booktitle={Proceedings of the Computer Vision and Pattern Recognition Conference},
  pages={2040--2049},
  year={2025}
}

@article{podell2023sdxl,
  title={Sdxl: Improving latent diffusion models for high-resolution image synthesis},
  author={Podell, Dustin and English, Zion and Lacey, Kyle and Blattmann, Andreas and Dockhorn, Tim and M{\"u}ller, Jonas and Penna, Joe and Rombach, Robin},
  journal={arXiv preprint arXiv:2307.01952},
  year={2023}
}

@inproceedings{wang2024microcinema,
  title={Microcinema: A divide-and-conquer approach for text-to-video generation},
  author={Wang, Yanhui and Bao, Jianmin and Weng, Wenming and Feng, Ruoyu and Yin, Dacheng and Yang, Tao and Zhang, Jingxu and Dai, Qi and Zhao, Zhiyuan and Wang, Chunyu and others},
  booktitle={Proceedings of the IEEE/CVF Conference on Computer Vision and Pattern Recognition},
  pages={8414--8424},
  year={2024}
}

@inproceedings{weng2024art,
  title={Art-v: Auto-regressive text-to-video generation with diffusion models},
  author={Weng, Wenming and Feng, Ruoyu and Wang, Yanhui and Dai, Qi and Wang, Chunyu and Yin, Dacheng and Zhao, Zhiyuan and Qiu, Kai and Bao, Jianmin and Yuan, Yuhui and others},
  booktitle={Proceedings of the IEEE/CVF Conference on Computer Vision and Pattern Recognition},
  pages={7395--7405},
  year={2024}
}

@inproceedings{xing2024simda,
  title={Simda: Simple diffusion adapter for efficient video generation},
  author={Xing, Zhen and Dai, Qi and Hu, Han and Wu, Zuxuan and Jiang, Yu-Gang},
  booktitle={Proceedings of the IEEE/CVF conference on computer vision and pattern recognition},
  pages={7827--7839},
  year={2024}
}

@article{ho2022imagen,
  title={Imagen video: High definition video generation with diffusion models},
  author={Ho, Jonathan and Chan, William and Saharia, Chitwan and Whang, Jay and Gao, Ruiqi and Gritsenko, Alexey and Kingma, Diederik P and Poole, Ben and Norouzi, Mohammad and Fleet, David J and others},
  journal={arXiv preprint arXiv:2210.02303},
  year={2022}
}

@article{blattmann2023stable,
  title={Stable video diffusion: Scaling latent video diffusion models to large datasets},
  author={Blattmann, Andreas and Dockhorn, Tim and Kulal, Sumith and Mendelevitch, Daniel and Kilian, Maciej and Lorenz, Dominik and Levi, Yam and English, Zion and Voleti, Vikram and Letts, Adam and others},
  journal={arXiv preprint arXiv:2311.15127},
  year={2023}
}

@article{wang2023videocomposer,
  title={Videocomposer: Compositional video synthesis with motion controllability},
  author={Wang, Xiang and Yuan, Hangjie and Zhang, Shiwei and Chen, Dayou and Wang, Jiuniu and Zhang, Yingya and Shen, Yujun and Zhao, Deli and Zhou, Jingren},
  journal={Advances in Neural Information Processing Systems},
  volume={36},
  pages={7594--7611},
  year={2023}
}

@inproceedings{wei2024dreamvideo,
  title={Dreamvideo: Composing your dream videos with customized subject and motion},
  author={Wei, Yujie and Zhang, Shiwei and Qing, Zhiwu and Yuan, Hangjie and Liu, Zhiheng and Liu, Yu and Zhang, Yingya and Zhou, Jingren and Shan, Hongming},
  booktitle={Proceedings of the IEEE/CVF Conference on Computer Vision and Pattern Recognition},
  pages={6537--6549},
  year={2024}
}

@inproceedings{wu2023tune,
  title={Tune-a-video: One-shot tuning of image diffusion models for text-to-video generation},
  author={Wu, Jay Zhangjie and Ge, Yixiao and Wang, Xintao and Lei, Stan Weixian and Gu, Yuchao and Shi, Yufei and Hsu, Wynne and Shan, Ying and Qie, Xiaohu and Shou, Mike Zheng},
  booktitle={Proceedings of the IEEE/CVF international conference on computer vision},
  pages={7623--7633},
  year={2023}
}

@inproceedings{xing2024dynamicrafter,
  title={Dynamicrafter: Animating open-domain images with video diffusion priors},
  author={Xing, Jinbo and Xia, Menghan and Zhang, Yong and Chen, Haoxin and Yu, Wangbo and Liu, Hanyuan and Liu, Gongye and Wang, Xintao and Shan, Ying and Wong, Tien-Tsin},
  booktitle={European Conference on Computer Vision},
  pages={399--417},
  year={2024},
  organization={Springer}
}

@inproceedings{zeng2024make,
  title={Make pixels dance: High-dynamic video generation},
  author={Zeng, Yan and Wei, Guoqiang and Zheng, Jiani and Zou, Jiaxin and Wei, Yang and Zhang, Yuchen and Li, Hang},
  booktitle={Proceedings of the IEEE/CVF Conference on Computer Vision and Pattern Recognition},
  pages={8850--8860},
  year={2024}
}

@article{brooks2024video,
  title={Video generation models as world simulators},
  author={Brooks, Tim and Peebles, Bill and Holmes, Connor and DePue, Will and Guo, Yufei and Jing, Li and Schnurr, David and Taylor, Joe and Luhman, Troy and Luhman, Eric and others},
  journal={OpenAI Blog},
  volume={1},
  number={8},
  pages={1},
  year={2024}
}

@article{lu2022dpm,
  title={Dpm-solver: A fast ode solver for diffusion probabilistic model sampling in around 10 steps},
  author={Lu, Cheng and Zhou, Yuhao and Bao, Fan and Chen, Jianfei and Li, Chongxuan and Zhu, Jun},
  journal={Advances in neural information processing systems},
  volume={35},
  pages={5775--5787},
  year={2022}
}

@inproceedings{zheng2023fast,
  title={Fast sampling of diffusion models via operator learning},
  author={Zheng, Hongkai and Nie, Weili and Vahdat, Arash and Azizzadenesheli, Kamyar and Anandkumar, Anima},
  booktitle={International conference on machine learning},
  pages={42390--42402},
  year={2023},
  organization={PMLR}
}

@inproceedings{meng2023distillation,
  title={On distillation of guided diffusion models},
  author={Meng, Chenlin and Rombach, Robin and Gao, Ruiqi and Kingma, Diederik and Ermon, Stefano and Ho, Jonathan and Salimans, Tim},
  booktitle={Proceedings of the IEEE/CVF conference on computer vision and pattern recognition},
  pages={14297--14306},
  year={2023}
}

@article{salimans2022progressive,
  title={Progressive distillation for fast sampling of diffusion models},
  author={Salimans, Tim and Ho, Jonathan},
  journal={arXiv preprint arXiv:2202.00512},
  year={2022}
}

@inproceedings{sauer2024adversarial,
  title={Adversarial diffusion distillation},
  author={Sauer, Axel and Lorenz, Dominik and Blattmann, Andreas and Rombach, Robin},
  booktitle={European Conference on Computer Vision},
  pages={87--103},
  year={2024},
  organization={Springer}
}

@article{luo2023lcm,
  title={Lcm-lora: A universal stable-diffusion acceleration module},
  author={Luo, Simian and Tan, Yiqin and Patil, Suraj and Gu, Daniel and von Platen, Patrick and Passos, Apolin{\'a}rio and Huang, Longbo and Li, Jian and Zhao, Hang},
  journal={arXiv preprint arXiv:2311.05556},
  year={2023}
}

@article{lu2024simplifying,
  title={Simplifying, stabilizing and scaling continuous-time consistency models},
  author={Lu, Cheng and Song, Yang},
  journal={arXiv preprint arXiv:2410.11081},
  year={2024}
}

@article{geng2024consistency,
  title={Consistency models made easy},
  author={Geng, Zhengyang and Pokle, Ashwini and Luo, William and Lin, Justin and Kolter, J Zico},
  journal={arXiv preprint arXiv:2406.14548},
  year={2024}
}

@article{song2020score,
  title={Score-based generative modeling through stochastic differential equations},
  author={Song, Yang and Sohl-Dickstein, Jascha and Kingma, Diederik P and Kumar, Abhishek and Ermon, Stefano and Poole, Ben},
  journal={arXiv preprint arXiv:2011.13456},
  year={2020}
}

@article{gu2024coherent,
  title={Coherent Video Inpainting Using Optical Flow-Guided Efficient Diffusion},
  author={Gu, Bohai and Luo, Hao and Guo, Song and Dong, Peiran and Zhou, Qihua},
  journal={arXiv preprint arXiv:2412.00857},
  year={2024}
}

@article{zi2025minimax,
  title={Minimax-remover: Taming bad noise helps video object removal},
  author={Zi, Bojia and Peng, Weixuan and Qi, Xianbiao and Wang, Jianan and Zhao, Shihao and Xiao, Rong and Wong, Kam-Fai},
  journal={arXiv preprint arXiv:2505.24873},
  year={2025}
}

@article{liu2026understanding,
  title={From Understanding to Erasing: Towards Complete and Stable Video Object Removal},
  author={Liu, Dingming and Wang, Wenjing and Li, Chen and Lyu, Jing},
  journal={arXiv preprint arXiv:2604.01693},
  year={2026}
}

@inproceedings{xu2018youtube,
  title={Youtube-vos: Sequence-to-sequence video object segmentation},
  author={Xu, Ning and Yang, Linjie and Fan, Yuchen and Yang, Jianchao and Yue, Dingcheng and Liang, Yuchen and Price, Brian and Cohen, Scott and Huang, Thomas},
  booktitle={Proceedings of the European conference on computer vision (ECCV)},
  pages={585--601},
  year={2018}
}

@inproceedings{perazzi2016benchmark,
  title={A benchmark dataset and evaluation methodology for video object segmentation},
  author={Perazzi, Federico and Pont-Tuset, Jordi and McWilliams, Brian and Van Gool, Luc and Gross, Markus and Sorkine-Hornung, Alexander},
  booktitle={Proceedings of the IEEE conference on computer vision and pattern recognition},
  pages={724--732},
  year={2016}
}

@inproceedings{lai2018learning,
  title={Learning blind video temporal consistency},
  author={Lai, Wei-Sheng and Huang, Jia-Bin and Wang, Oliver and Shechtman, Eli and Yumer, Ersin and Yang, Ming-Hsuan},
  booktitle={Proceedings of the European conference on computer vision (ECCV)},
  pages={170--185},
  year={2018}
}

\end{document}